\documentclass[journal]{IEEEtran}

\usepackage{color}
\usepackage{transparent}
\usepackage{import}
\usepackage{multirow,bigdelim}
\usepackage{balance}
\usepackage{graphicx,xcolor}
\usepackage{psfrag}
\usepackage{amsmath,amsfonts} 
\usepackage{gradient-text}

\usepackage{amssymb}  
\usepackage{makecell}

\usepackage{booktabs,tabularx,caption,ragged2e}
\usepackage{cite}
\usepackage{times,xspace}
\usepackage[font={footnotesize}]{caption}
\usepackage{makecell}
\usepackage{algorithm} 
\usepackage{algpseudocode} 
\usepackage{nicefrac}
\usepackage{romannum}
\usepackage{subcaption}
\usepackage{siunitx}
\usepackage{url}
\usepackage[bottom]{footmisc}

\usepackage[makeroom]{cancel}

\usepackage{mathtools}

\usepackage{array}
\usepackage{tcolorbox} 

\usepackage{mdframed} 
\usepackage{stmaryrd}
\usepackage{svg}
\usepackage{accents}
\usepackage{ntheorem}
\theoremstyle{break}
\theoremheaderfont{\normalfont\bfseries}
\theoremseparator{:}

\newcommand{\highlight}[1]{%
	\raisebox{0pt}[0pt][0pt]{\colorbox{TUMyellow!20}{\smash{#1}}}%
}

\usepackage{tikz} 
\makeatletter

\usepackage[bottom]{footmisc}

\usepackage{pifont}
\newcommand{\cmark}{\ding{51}}%
\newcommand{\xmark}{\ding{55}}%

\definecolor{gridcolor}{RGB}{220,220,220}
\definecolor{lightgray}{RGB}{230,230,230}
\definecolor{darkergray}{RGB}{100,100,100}
\definecolor{background}{RGB}{239,239,239}
\definecolor{reachgray}{RGB}{180,180,180}
\definecolor{reachgraylight}{RGB}{144,144,144}

\definecolor{RPTHred}{RGB}{227, 27, 35}
\definecolor{TUMblue}{RGB}{0, 101, 189}
\definecolor{TUMyellow}{RGB}{254, 215, 2}
\definecolor{TUMblue1}{RGB}{154, 188, 228}
\definecolor{TUMblue2}{RGB}{194, 215, 239}
\definecolor{TUMblue3}{RGB}{215, 228, 244}
\definecolor{TUMblue4}{RGB}{227, 238, 250}
\usepackage{csquotes}	
\makeatother
\newtcolorbox{mytransparentbox}[1][]{%
	opacityback=0.65,
	opacityframe=0.0,
	colback=white,
	left=0pt,
	right=0pt,
	top=0pt,
	bottom=0pt,
	boxrule=0pt,
	halign=center,
	valign=center,
	#1
}
\usepackage{hyperref}
\algrenewcommand\algorithmicrequire{\textbf{Input:}}
\algrenewcommand\algorithmicensure{\textbf{Output:}}
\algnewcommand\algorithmicinput{\textbf{Optional Input:}}
\algnewcommand\INPUT{\item[\algorithmicinput]}
\newif\ifdraft
\drafttrue  

\ifdraft
\renewcommand{\textcolor}[2]{#2}  
\else
\fi

\usepackage{tikz} 
\newcommand\copyrighttext{%
	\footnotesize \copyright 2026 IEEE. Personal use of this material is permitted. Permission from IEEE must be obtained for all other uses, in any current or future media, including reprinting/republishing this material for advertising or promotional purposes, creating new collective works, for resale or redistribution to servers or lists, or reuse of any copyrighted component of this work in other works.}
\newcommand\copyrightnotice{%
	\begin{tikzpicture}[remember picture,overlay]
		\node[anchor=south,yshift=5pt] at (current page.south) {\fbox{\parbox{\dimexpr\textwidth-\fboxsep-\fboxrule\relax}{\copyrighttext}}};
	\end{tikzpicture}%
}
\begin{document}

\title{\textsc{\gradientRGB{SanDRA}{0, 101, 189}{254, 215, 2}}: Safe Large-Language-Model-Based Decision Making for Automated Vehicles Using Reachability Analysis}

\author{Yuanfei Lin$^*$, Sebastian Illing$^*$, and Matthias Althoff%
		\thanks{$^*$ The first two authors have contributed equally to this work.}
	\thanks{
	Manuscript received Month Date, Year; revised Month Date, Year.
	}
\thanks{The authors are with the Department of Computer Engineering, Technical University of Munich, 85748 Garching, Germany. Matthias Althoff is also with the Munich Center for Machine Learning (MCML), 80538 Munich, Germany. {\tt\small \{yuanfei.lin, sebastian.illing, althoff\}@tum.de} \textit{ (Corresponding author: Yuanfei Lin.)} }

}

\markboth{Journal of XX,~Vol.~XX, No.~X, Month~Year}%
{Shell \MakeLowercase{\textit{et al.}}: Bare Demo of IEEEtran.cls for IEEE Journals}

\maketitle

\begin{abstract}
		Large language models \textcolor{red}{(LLMs)} have been widely applied to knowledge-driven decision-making for automated vehicles due to their strong generalization and reasoning capabilities. However, the safety of the resulting decisions cannot be ensured due to possible hallucinations and the lack of integrated vehicle dynamics.
To address this issue, we propose \textsc{SanDRA}, the first \underline{sa}fe {la}rge-{la}\underline{n}guage-model-based \underline{d}ecision making framework for automated vehicles using \underline{r}eachability \underline{a}nalysis. Our approach starts with a comprehensive description of the driving scenario to prompt \textcolor{red}{LLMs} to generate and rank feasible driving actions. These actions are translated into temporal logic formulas that incorporate formalized traffic rules, and are subsequently integrated into reachability analysis to eliminate unsafe actions. We validate our approach in both open-loop and closed-loop driving environments using off-the-shelf and finetuned \textcolor{red}{LLMs}, showing that it can provide provably safe and, where possible, legally compliant driving actions, even under high-density traffic conditions. 
To ensure transparency and facilitate future research, all code and experimental setups are publicly available at \href{https://github.com/CommonRoad/SanDRA}{github.com/CommonRoad/SanDRA}. 
\end{abstract}


\IEEEpeerreviewmaketitle
\section{Introduction}\label{sec:introduction}
\begin{figure}[!t]%
	\centering
	\vspace{2.5mm}
	\def\svgwidth{0.98\columnwidth}\footnotesize
	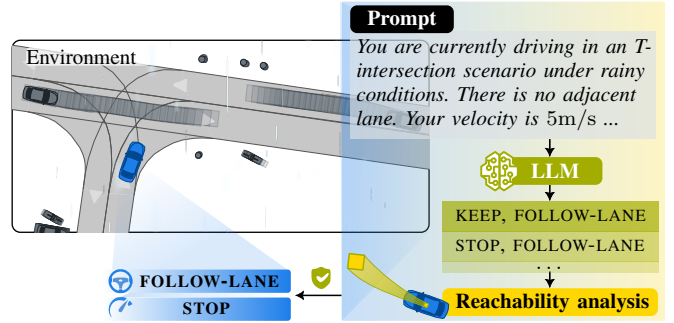
	\caption{An example usage of \textsc{SanDRA}, where the LLM is prompted to generate a ranked list of longitudinal and lateral action pairs ordered from best to worst. The action pair corresponding to stopping while staying on the current lane is verified as safe using reachability analysis and then executed. 
	} \label{fig:schema_general}

\end{figure}

\IEEEPARstart{D}{ecision}-making is essential for automated vehicles to navigate safely and interact appropriately with other traffic participants. Traditional methods rely on finite state machines for simple scenarios \cite[Sec. II-B]{Paden2016} and neural networks to handle more complex behaviors like lane changes \cite{Krasowski2020a}. Recent efforts have shifted toward a knowledge-driven paradigm with the emergence of large language models (LLMs) that enable human-like and context-sensitive decision-making \cite{cui2024survey, wang2024survey}. \copyrightnotice 
However, their limited understanding of underlying physical models, tendency to hallucinate, and implicit consideration of traffic rules pose critical safety concerns that are unacceptable for real-world deployment. To address this, we propose using reachability analysis to verify the safety of decisions generated by LLMs before their execution (cf. Fig.~\ref{fig:schema_general}).



\subsection{Related Work}\label{sec:li_ov}
Below, we concisely review related works on LLM-based decision-making and safety verification approaches for automated vehicles.
\paragraph {\it LLMs as Decision Makers}
	LLMs are trained on vast and diverse internet-scale data, enabling them to understand common driving logs and reason about traffic scenarios \cite{cui2024survey, wang2024survey}. 
For this reason, they have been applied to automated driving to generate software programs, high-level actions, waypoints, and control signals; an overview is shown in Tab.~\ref{tab:safety}. The majority of these approaches employ LLMs to choose suitable driving actions given sensor inputs and/or narrative descriptions of the driving context. However, the safety of the selected actions is typically unverified or delegated to hardcoded rules \cite{jin2024surrealdriver, cai2024driving}, another LLM \cite{wen2023dilu, shinn2023reflexion, jiang2024koma}, or software modules evaluating criticality measures \cite{zhou2024safedrive, pang2024large, kou2025padriver}.  
Despite being instructed to generate safe actions,  LLMs often fail to fully capture critical scenario information and may hallucinate, leading to frequent rule violations and collisions \cite{wen2023dilu, wang2023drivemlm, kou2025padriver}. A similar fragility arises from their limited understanding of vehicle dynamics when directly generating waypoints to be followed, resulting in kinematically infeasible trajectories \cite{mao2023gpt, sima2023drivelm, shao2023lmdrive}. In contrast, another category of work employs LLMs to adjust key components of the software stack without altering its overall structure \cite{drplanner, baumann2025enhancing, sha2025languagempc}, allowing existing safety mechanisms to remain effective. However, these approaches still utilize only a limited portion of the full potential of LLMs and lack formal guarantees of correctness.
\renewcommand{\arraystretch}{1.05}
\begin{table}[t!]
	\begin{center}
		\vspace{1.5mm}
		\caption{Related work on LLMs as decision makers and their support for safety-verified decision outputs.} 
		\label{tab:safety}
		\begin{tabular}{@{}p{1.4cm}p{4cm}cc@{}}
			\toprule[1.pt]
			\textbf{Work} &  \textbf{Output} & \makecell[l]{\textbf{Safety}\\ \textbf{check}} &  \makecell[l]{\textbf{Formally}\\ \textbf{correct}}\\\midrule
			\cite{wen2023dilu, shinn2023reflexion, jiang2024koma, cai2024driving, jin2024surrealdriver, zhou2024safedrive} &  High-level actions  & \cmark  & \xmark \\ 
			\cite{kou2025padriver, wang2023drivemlm, pang2024large} & High-level actions  & \xmark &\xmark \\
			\cite{mao2023gpt, sima2023drivelm} & High-level actions \& waypoints  & \xmark &  \xmark \\
			\cite{sha2025languagempc, baumann2025enhancing} & High-level actions \& controller parameters &  \cmark & \xmark \\
			\cite{shao2023lmdrive} & Waypoints & \xmark & \xmark\\
			\cite{drplanner, cui2024personalized}& Programs & \cmark & \xmark \\ 
			\cite{chen2023driving, xu2024drivegpt4} & Control signals & \xmark & \xmark\\
			\cite{cui2024drive, han2025traffic} & Traffic descriptions & \xmark & \xmark \\
			\cite{hwang2024emma} &  Traffic descriptions \& waypoints & \xmark & \xmark \\  
			\midrule
			\textbf{Ours} & \textbf{High-level actions \& trajectories} & \cmark & \cmark \\
			
			\bottomrule
		\end{tabular}
	\end{center}
\end{table}

\paragraph {\it Safety Verification}
Formal verification is the process of verifying whether system executions comply with a formal specification.		
For a review on verifying motion plans for automated vehicles, we refer to \cite{mehdipour2023formal}. Among the various techniques, \textcolor{red}{such as backup control barrier functions \cite{chen2021backup}, gatekeeper \cite{agrawal2024gatekeeper}, and shielding \cite{alshiekh2018safe}}, reachability analysis is the most widely used approach for online safety verification of automated vehicles \cite{althoff2014online, christian2020nature, ahn2020reachability, nomoretraffic}. The specification is typically expressed as a temporal logic formula, such as linear temporal logic (LTL) or metric temporal logic (MTL), encompassing safety specifications \cite{Arechiga2019, hekmatnejad2019encoding}, formalized traffic rules \cite{Esterle2020, Maierhofer2020a, Maierhofer2022a}, and driving actions \cite{corso2020interpretable, sahin2020autonomous, klischat2020synthesizing}. 
These formulas can be integrated into the software stack of automated vehicles to generate specification-compliant trajectories, monitor traffic rule compliance, and repair trajectories that violate these specifications \cite{plaku2016motion, nomoretraffic, yuanfei2024}. 
Specifically, by coupling reachability analysis with model checking, one can obtain driving corridors for automated vehicles that satisfy the given formulas \cite{liu2023specification, lercher2024specification}. If the reachable set defining these corridors becomes empty within the planning horizon, a specification-compliant action no longer exists. In addition, verification techniques can be integrated into deep learning-based frameworks as safety layers, such as reinforcement learning \cite{li2019formal, Krasowski2020a, krasowski2024provable}. 
However, to the best of our knowledge, no such integration exists yet for an LLM-based decision-making framework. 


\subsection{Contributions}\label{sec:contri}
In this work, we present \textsc{SanDRA}, the first framework for integrating LLMs safely into decision making for automated vehicles through reachability analysis, combining the strengths of machine learning and formal methods. In particular, our contributions are:
\begin{enumerate}
	\item designing a lightweight prompt that guides LLMs to generate and prioritize driving action candidates;
	\item mapping driving actions to LTL formulas and using them to label human trajectories for LLM finetuning; 
	\item combining reachability analysis with the LTL formulas of action candidates, along with formalized traffic rules, to verify their safety. The computed reachable sets can be used as constraints for the downstream trajectory planning module; and
	\item evaluating safety performance in both open- and closed-loop simulations, with the closed-loop setup incorporating both most-likely and set-based predictions of surrounding traffic participants.
\end{enumerate}

The remainder of this article is structured as follows: 
Sec.~\ref{sec:pre} describes the general setup and necessary preliminaries. Sec.~\ref{sec:sandra} details the proposed framework for safe LLM-based decision making. In Sec.~\ref{sec:eva}, we demonstrate the advantages of our approach. Finally, Sec.~\ref{sec:conc} concludes the article.

\section{Preliminaries}\label{sec:pre}
\subsection{General Setup}
The vehicle for which the decision-making is performed is referred to as the \textit{ego vehicle}. We use the index $k \in \mathbb{N}_0$ to denote the discrete time step corresponding to the continuous time $t_k = k\Delta t$, where $\Delta t \in \mathbb{R}_{+}$ is a fixed time increment. The dynamics of the ego vehicle is modeled as:
\begin{equation}\label{eq:dynamics}
	\boldsymbol{x}_{k+1} = f(\boldsymbol{x}_k, \boldsymbol{u}_k),
\end{equation}
where $\boldsymbol{x}_k\in\mathcal{X}_k$ is the state within the permissible state space $\mathcal{X}_k\in\mathbb{R}^{n_x}$, and $\boldsymbol{u}_k\in\mathcal{U}_k$ is the control input within the admissible input space $\mathcal{U}_k\in\mathbb{R}^{n_u}$. Without loss of generality, we set the current time step to $0$ and denote the decision-making horizon by $h$. The solution of (\ref{eq:dynamics}) at time step $k$ is denoted as $\chi_k(\boldsymbol{x}_0, \boldsymbol{u}_{[0,k]})$, where $\boldsymbol{u}_{[0,k]}$ is the input trajectory. The solution over the time interval $[0, k]$ is denoted by $\chi_{[0, k]}(\boldsymbol{x}_0, \boldsymbol{u}_{[0,k]})$. 
The operator $\mathrm{occ}(\cdot)$ returns the spatial occupancy of the entity. 

\begin{figure}[!t]%
	\centering
	\vspace{.5mm}
	\def\svgwidth{1\columnwidth}\footnotesize
	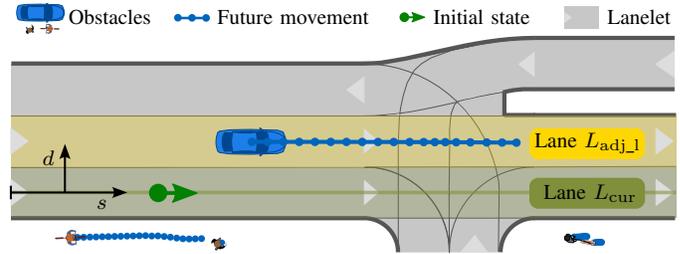
	\caption{Exemplary road network with lanes defined by lanelets. The ego vehicle is currently following lane $L_{\mathrm{cur}}$, as determined by the high-level routing module to reach the goal. Alternatively, it may switch to the left-adjacent lane $L_{\mathrm{adj\_l}}$
		instead of remaining on $L_\mathrm{cur}$.
	}\label{fig:cr_scenario}
\end{figure}

As shown in Fig.~\ref{fig:cr_scenario}, we describe the road network as a set of lanelets $l_{i}$, $i \in\mathbb{N}_{+}$, each defined by left and right boundaries represented as polylines  \cite{Bender2014}. 
 We then define a lane $L_j$ as a set of connected lanelet sections. 
   A lane $L_j$ is adjacent to another lane $L_{j'}$ if there exists a lanelet in $L_j$ whose border fully overlaps with the border of a lanelet in  $L_{j'}$.
We denote the lane currently occupied by the ego vehicle as $L_{\mathrm{cur}}$, with ${L}_{\mathrm{adj\_l}}$ and ${L}_{\mathrm{adj\_r}}$ referring to its left and right adjacent lanes, respectively.
Using the centerline of $L_{\mathrm{cur}}$, we can construct a curvilinear coordinate system in which vehicles are described by their longitudinal position $s$ and lateral deviation $d$ \cite{Gerald2024} (cf. Fig.~\ref{fig:cr_scenario}). In addition, we denote the relative longitudinal distance, orientation, velocity, steering angle, and acceleration as $\Delta s$, $\theta$, $v$, $\delta$, and $a$, respectively.
\subsection{Specification-Compliant Reachability Analysis}
\begin{figure*}[!t]%
	\centering
	\vspace{2.5mm}
	\def\svgwidth{1.96\columnwidth}\footnotesize
	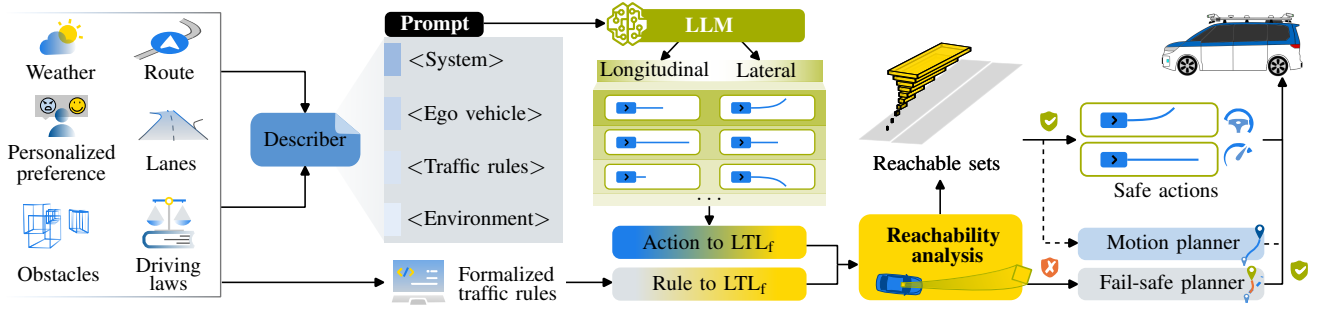
	\caption{\textsc{SanDRA} overview diagram. Our tool \textsc{SanDRA} takes planning and environmental information as inputs and processes them to generate a structured description of the current scenario. The description is then used to prompt the LLM to produce a ranked sequence of longitudinal and lateral action pairs, ordered from best to worst. 
		After converting the actions and traffic rules to LTL$_\text{f}$ formulas, we take their conjunction and apply reachability analysis to verify the safety of the resulting behavior. Once verified, the action pair is either executed directly or its corresponding reachable sets are passed to the motion planner to generate safe trajectories. If no verified actions are available, a fail-safe plan is executed.} \label{fig:pipeline}\vspace{0mm}
\end{figure*}%
Since decision making typically focuses on short-term driving behaviors, we employ reachability analysis over a finite time horizon and formalize driving actions as formulas in LTL over finite traces (LTL$_{\text{f}}$) \cite{de2013linear}. In this setting, temporal operators are interpreted over finite traces bounded by the decision-making horizon $h$. To ensure safety over an infinite time horizon, fail-safe planning is required \cite[Sec.~X]{nomoretraffic}. 
\subsubsection{Linear Temporal Logic over Finite Traces}
Given a set of atomic propositions $\mathcal{AP}$ and formulas $\varphi$, $\varphi_1$, and $\varphi_2$, the syntax of LTL$_{\text{f}}$ is defined according to the grammar \cite[Sec.~2]{de2013linear}:
\begin{equation}\label{eq:ltlf}
	\varphi \Coloneqq \sigma \ |\ \lnot \varphi \ |\ \varphi_1 \land \varphi_2 \ |\ \mathbf{G} \varphi \ |\ \mathbf{F} 
	\varphi,
\end{equation} 
where $\sigma\in\mathcal{AP}$ is an atomic proposition, and the symbols $\lnot$ and $\land$ denote Boolean \textit{negation} and \textit{conjunction} connectives, respectively. The temporal operator $\mathbf{G}\varphi$ requires that $\varphi$	will \textit{always} hold over the finite trace, whereas $\mathbf{F}\varphi$ requires that $\varphi$ \textit{eventually} holds within the finite trace. 
Further logic operators can be derived from (\ref{eq:ltlf}) (see \cite[Sec.~5.1.1]{baier2008principles}), such as $\varphi_1 \lor \varphi_2 \coloneqq\lnot (\lnot\varphi_1\land\lnot\varphi_2)$ (\textit{disjunction}) and $\varphi_1 \rightarrow \varphi_2\coloneqq\lnot \varphi_1\lor \varphi_2$ (\textit{implication}).
If a finite trace  $\chi_{[0, k]}(\boldsymbol{x}_0, \boldsymbol{u}_{[0,k]})$ satisfies the specification $\varphi$, we write $\chi_{[0, k]}(\boldsymbol{x}_0, \boldsymbol{u}_{[0,k]})\models\varphi$. For formalized driving actions and traffic rules, each atomic proposition is typically associated with a predicate, defined as a function of vehicle states. 
\subsubsection{Specification-Compliant Reachable Sets}
We assume that the predicted time-varying occupancies of all obstacles are given as $\mathcal{O}_k\subset \mathbb{R}^2$ at each time step $k$. Based on this, we define the set of forbidden states of the ego vehicle as $\mathcal{X}_k^{\mathtt{F}} = \{\boldsymbol{x}_k\in\mathcal{X}_k \ | \ \mathrm{occ}(\boldsymbol{x}_k) \cap \mathcal{O}_k \neq \emptyset\}$. The specification-compliant reachable set of the ego vehicle $\mathcal{R}_k^{\mathtt{e}}$ at time step~$k$ is defined as the set of states reachable from an initial state in~$\mathcal{X}_0$, while avoiding the forbidden states~$\mathcal{X}_\tau^{\mathtt{F}}$ for every step $\tau \in \{0, \dots, k\}$, via a drivable trajectory that satisfies~$\varphi$. Formally, we have \cite[Sec.~II-C]{lercher2024specification}:
\begin{equation}\label{eq:reach}
	\begin{split}
	\!\!\mathcal{R}_k^{\mathtt{e}} (\varphi)\coloneqq\!\! \bigg\{
\chi_{k}(\boldsymbol{x}_0, & \boldsymbol{u}_{[0,k]}) \bigg| \exists \boldsymbol{x}_0\in\mathcal{X}_0, \forall \tau \in\{0, \dots, k\}, \\
&\exists\boldsymbol{u}_\tau \in \mathcal{U}_\tau: \chi_{\tau}(\boldsymbol{x}_0, \boldsymbol{u}_{[0,\tau]}) \notin \mathcal{X}_k^{\mathtt{F}} \ \land\\
 & \qquad\qquad \quad \chi_{[0, k]}(\boldsymbol{x}_0, \boldsymbol{u}_{[0,k]})\models\varphi
\bigg\}.
	\end{split}
\end{equation}
Obtaining $\mathcal{R}_k^{\mathtt{e}}(\varphi)$ is generally infeasible~\cite{platzer2007image}. Therefore, we use a tight overapproximation $\mathcal{R}_k(\varphi) \supseteq \mathcal{R}_k^{\mathtt{e}}(\varphi)$ that encloses all safe and specification-compliant states \textcolor{red}{\cite{althoff2014online, lercher2024specification}}.

\subsection{Problem Formulation}
This article addresses two key challenges: designing lightweight scenario representations for prompting LLMs for decision making of automated vehicles, and bridging their natural language outputs with safety verification techniques. For the latter, we focus on translating language to LTL$_\text{f}$ formulas $\varphi$; we additionally verify the legal safety\footnote{\textcolor{red}{We define legal safety as the property that the ego vehicle cannot cause any traffic rule violations \cite{nomoretraffic}.}} of each candidate action by ensuring that its corresponding reachable set for the ego vehicle remains non-empty throughout the decision-making horizon $h$, i.e.,
\begin{equation*}
\forall k\in\{0,\dots,h\}: \mathcal{R}_k(\varphi)\neq\emptyset.
\end{equation*} 
We emphasize that the degree of safety guarantee depends on the accuracy of predictions for other traffic participants. If an inaccurate prediction leads to a safety-critical situation, we apply the safety concept described in \cite[Sec. X]{nomoretraffic}. 

\section{\textsc{SanDRA}}\label{sec:sandra}
This section first introduces the overall algorithm of \textsc{SanDRA} in Sec.~\ref{subsec:oa}, followed by detailed explanations of its key components in Secs.~\ref{subsec:prompt}-\ref{subsec:safety_v}.
\subsection{Overall Algorithm}\label{subsec:oa}
As shown in Fig.~\ref{fig:pipeline}, at each decision-making cycle, \textsc{SanDRA} takes as input both environmental and planning data, including high-level routes from navigation systems, drivable lanes, detected obstacles, and other contextual information such as traffic rules \cite{vienna_convention} and weather conditions.\footnote{\href{https://openweathermap.org/api}{https://openweathermap.org/api}} To enable personalization, different driving styles can be incorporated through predefined driving preferences \cite{kou2025padriver} or verbal commands from humans \cite{cui2024personalized}. 
\textcolor{red}{All input data are automatically converted into a narrative description within \textsc{SanDRA}.
This is achieved by first mapping each input type to predefined semantic categories, e.g., vehicle state, user command, traffic rules, second instantiating template sentences for each category, and finally assembling these sentences into a coherent story that preserves temporal and spatial relationships. The resulting narrative is used to prompt the LLM, which performs reasoning and generates a sequential ordering of $\kappa\in\mathbb{N}_{+}$ feasible longitudinal and lateral action pairs (cf. Tab.~\ref{tab:action} and \textcolor{red}{Sec.~\ref{subsec:prompt}}).}
Although the actions are presented separately for clarity, they are in fact considered jointly to generalize across diverse road structures, unlike the other approaches for generating driving decisions listed in Tab.~\ref{tab:safety}. 
Note that to obtain a comprehensive representation of the scenario, other sensor modalities can also be tokenized as vision tokens \cite{radford2021learning} and be combined with textual tokens from the narrative in a shared embedding space, enabling the use of multi-modal LLMs \cite{kou2025padriver}.
Afterwards, the returned action pairs are converted into LTL\textsubscript{f} formulas (cf. Sec.~\ref{subsec:action2ltl}), which are then conjuncted with the formalized traffic rules for legal safety (cf. Sec.~\ref{subsec:rules}). With this, the safety verification is carried out through the computation of reachable sets that satisfy the formulas (cf. Sec.~\ref{subsec:safety_v}).
 The first verified action pair can be executed by the controller, or its corresponding reachable sets can be used as a constraint by various trajectory planners to generate safe trajectories \cite[Sec.~VIII]{nomoretraffic}. When the connection to the cloud is lost or other failures occur that cause the LLM to fail to respond in time, produce hallucinations, or when no feasible verified action is available, we revert to a fail-safe trajectory to guarantee safety \cite[Sec.~X]{nomoretraffic} (cf. Sec.~\ref{subsec:safety_v}).







\subsection{Prompt Design and LLM Querying}\label{subsec:prompt}
We construct a describer module that converts driving context information into \textcolor{red}{template-based} descriptive text, structured into three key components, with full details provided in Fig.~\ref{fig:prompt}; the latter two form the user prompt. In addition to the automatically converted content, minimal manual inputs specify driving styles, hyperparameters, and regulation-related aspects. 
 The modular design of the prompt enables easy adaptation and extension. \textcolor{red}{Note that all directional aspects, such as {\{left / right\}} lanes or {\{same / opposite\}}  directions, are derived from the topological relations of the road network (cf. Fig.~\ref{fig:commonroad_sce}).}
\subsubsection{System Prompt} In the system prompt, we introduce the decision-making task for LLMs, 
which may also include personalized commands, e.g., ``\textit{drive faster}”. 
The LLM is guided to generate longitudinal and lateral action pairs ranked from best to worst by prompting it to decide in a chain-of-thought manner \cite{wei2022chain}. Specifically, we prefilter and provide the feasible actions based on current lane information, generalizing the approaches from \cite{wen2023dilu, jin2024surrealdriver, kou2025padriver}. The longitudinal actions are determined by whether the maneuvers are allowed in the current driving context, whereas the lateral actions are obtained by querying the lane structure at the current position of the ego vehicle (cf. Fig.~\ref{fig:cr_scenario}). For instance, in Fig.~\ref{fig:schema_general}, all longitudinal actions are contextually allowed, since the scenario is set in a rural environment; however, as the occupied lane of the ego vehicle has no adjacent lanes, the only feasible lateral action is \textsc{follow-lane}. 

\begin{figure}\small
	\vspace{2mm}
	\begin{mdframed}[backgroundcolor=TUMblue1,   innerleftmargin=10pt,   
		innerrightmargin=10pt, linecolor=white,      
		linewidth=0pt]
		\textbf{{\textless System\textgreater}}: You are driving a car and need to make a high-level driving decision.  \highlight{[User command]}.  First, carefully observe the environment; then, reason through your decision step by step and present it in natural language, and finally, return the top \highlight{[$\kappa$]} advisable longitudinal–lateral action pairs ranked from best to worst. \highlight{\{Feasible longitudinal actions\}}.  \highlight{\{Feasible lateral actions\}}. The past action pairs are \highlight{\{previous actions\}}.
	\end{mdframed}
	\hfill
	\vspace{-4mm}
	\begin{mdframed}[backgroundcolor=TUMblue2,   innerleftmargin=10pt,   
		innerrightmargin=10pt, linecolor=white,      
		linewidth=0pt]
		\textbf{{\textless Ego vehicle\textgreater}}: You are currently driving in a \highlight{\{road type\}} scenario with \highlight{\{weather condition\}}. (There is a \highlight{\{left / right\}}-adjacent lane  with the \highlight{\{same / opposite\}} direction. There are incoming lanes on the \highlight{\{left / right\}}.) Your velocity is \highlight{\{$v$\}}, your orientation is \highlight{\{$\theta$\}}, your steering angle is \highlight{\{$\delta$\}}, and your acceleration is \highlight{\{$a$\}}. 
	\end{mdframed}
	\hfill
		\vspace{-4mm}
		\begin{mdframed}[backgroundcolor=TUMblue4,   innerleftmargin=10pt,   
		innerrightmargin=10pt, linecolor=white,      
		linewidth=0pt]
		\textbf{{\textless Obstacles\textgreater}}: Here is an overview of all relevant obstacles surrounding you: 
		
		\highlight{\{Obstacle 1: \{Lane information\}. \{State\}. \{(Criticality)\}.\}}. 
		\highlight{\{Obstacle 2: \{Lane information\}. \{State\}. \{(Criticality)\}.\}}. $\dots$
	\end{mdframed}	\hfill
	\vspace{-4mm}
	\caption{LLM prompt design. Automatically generated content in the prompt is marked with \highlight{$\{$curly brackets$\}$}, while manual inputs are in \highlight{$[$square} \highlight{brackets$]$}, and \highlight{$($round brackets$)$} denote optional elements depending on the road structure and its regulatory features.}\label{fig:prompt}	\vspace{-2mm}
\end{figure}

\subsubsection{Ego Vehicle Prompt} This section provides dynamic information about the ego vehicle, including environmental context and road network configurations. It also covers the current vehicle state, including velocity, orientation, steering angle, and acceleration. A snippet of the ego vehicle prompt is shown in Fig.~\ref{fig:schema_general}.  
\subsubsection{Obstacle Prompt} This prompt block describes the surrounding traffic participants in the scenario, including their IDs, types, and states. These obstacles are identified either through lane-based adjacency or using Lidar-like beams relative to the ego vehicle \cite[Fig.~2]{commonroad-rl}. In addition, the description is enriched with criticality measures \cite{lin2023commonroad}, e.g., time-to-collision, to quantify the threat level posed by the obstacles to the ego vehicle. For instance, for dynamic vehicles at interstates, we use the following prompt:
		\begin{mdframed}[backgroundcolor=TUMblue4,   innerleftmargin=10pt,   
	innerrightmargin=10pt, linecolor=white,      
	linewidth=0pt]\small
	\textbf{Car \highlight{\{id\}}}: it is driving on your \highlight{\{same / left adjacent / right} \highlight{adjacent\}} lane in the \highlight{\{same / opposite\}} direction and is \highlight{\{$\Delta s$ in front of / behind\}} you. Its velocity is \highlight{\{$v$\}}, its orientation is \highlight{\{$\theta$\}}, its steering angle is \highlight{\{$\delta$\}}, and its acceleration is \highlight{\{$a$\}}. (The time-to-collision is \highlight{\{time-to-collision\}}.)
\end{mdframed}	

The prompt is fed into the LLM, which can be either an off-the-shelf model or a model finetuned on recorded driving datasets. 
To enhance reliability and facilitate parsing, we define the desired output as a schema, i.e., a sequence of longitudinal-lateral action pairs, and guide the LLM to produce actions in a format that conforms to this schema.\footnote{\href{https://openai.com/index/introducing-structured-outputs-in-the-api/}{https://openai.com/index/introducing-structured-outputs-in-the-api/}}
Moreover, for finetuning, we use recorded driving datasets and automatically label each scenario, using the prompt design in Fig.~\ref{fig:prompt} as input and the action pair of a randomly selected vehicle -- labeled according to the LTL$_\text{f}$ formula shown in Tab.~\ref{tab:action} -- as output. When $\kappa>1$ action pairs are required, the labeled action pair is used as the top-$1$ option, and 
\textcolor{red}{the remaining positions are filled with other feasible action pairs chosen at random.}
The LLM is finetuned based on the collected input-output pairs to adapt it for our decision-making tasks \cite{hu2022lora}. 

\begin{table}[t!]
	\begin{center}
		\vspace{1.5mm}
		\caption{High-level action space for the ego vehicle and the corresponding LTL$_{\text{f}}$ formula.}
		\label{tab:action}
		\begin{tabular}{@{}p{2.2cm}p{2.5cm}l@{}}
			\toprule[1.pt]
			\textbf{Action} &  \textbf{Direction} & \textbf{LTL$_{\text{f}}$ formula}\\\midrule
			\textsc{Keep} & \multirow{4}{*}{Longitudinal action} & $\mathbf{G}(\left|a\right|\leq a_{\mathrm{lim}})$\\
			\textsc{Accelerate} & & $\mathbf{G}(a > a_{\mathrm{lim}})$\\
			\textsc{Decelerate} & & $\mathbf{G}(a < -a_{\mathrm{lim}})$\\
			\textsc{Stop} \vspace{1mm}& & $\mathbf{F}\mathbf{G}\big(\mathrm{in\_standstill}(\boldsymbol{x})\big)$\\
						\textsc{Follow-lane} & \multirow{3}{*}{Lateral action}  & $\mathbf{G}\big(\mathrm{in\_lane}(\boldsymbol{x}, L_{\mathrm{cur}})\big)$ \\
			\textsc{Left-lane} & & $\mathbf{F}\mathbf{G}\big(\mathrm{in\_lane}(\boldsymbol{x}, L_{\mathrm{adj\_l}})\big)$ \\
			\textsc{Right-lane} & & $\mathbf{F}\mathbf{G}\big(\mathrm{in\_lane}(\boldsymbol{x}, L_{\mathrm{adj\_r}})\big)$ \\

			\bottomrule
		\end{tabular}\vspace{-3mm}
	\end{center}
\end{table}
\subsection{Action to LTL$_{\text{f}}$}\label{subsec:action2ltl}
   We incorporate string representations of actions into reachability analysis by converting them to LTL$_\text{f}$ formulas. To the best of our knowledge, such a conversion of driving actions has not yet been presented in the literature. Analogous to the traffic rule formalization in \cite{Maierhofer2020a, Maierhofer2022a}, we propose a two-step process: first, concretizing the string descriptions, and second, formalizing them using logic. 
   For instance, the action \textsc{stop} can be concretized as \enquote{\textit{the ego vehicle eventually maintaining a velocity close to zero within the decision-making horizon}}, where the closeness to zero is quantified by the measurement uncertainty $v_{\mathrm{err}}\in \mathbb{R}_{0}$. Afterward, we extract and specify predicates in higher-order logic based on the concretization, reusing existing ones from the formalized traffic rules \cite{Maierhofer2020a, Maierhofer2022a} where applicable. With this, the action \textsc{stop} is formalized as
	$\mathbf{F}\mathbf{G}\big(\mathrm{in\_standstill}(\boldsymbol{x})\big)$,
where the predicate $\mathrm{in\_standstill}(\boldsymbol{x})\leftrightarrow - v_{\mathrm{err}} \leq v \leq v_{\mathrm{err}}$ indicates whether the ego vehicle is in standstill \cite[Sec.~IV-B]{Maierhofer2022a}.

As summarized in Tab.~\ref{tab:action}, we formalize the most commonly used longitudinal and lateral actions based on their temporal and spatial concretizations, which can be easily extended to additional action definitions. The longitudinal actions \textsc{keep}, \textsc{accelerate}, and \textsc{decelerate} are interpreted as confining the acceleration of the ego vehicle to a certain range and are distinguished based on a predefined acceleration threshold $a_{\mathrm{lim}} \in \mathbb{R}_{0}$. 
In contrast, lateral actions are based on the predicate $\mathrm{in\_lane}(\boldsymbol{x}, L_j)	\leftrightarrow \bigvee_{l_i\in L_j} \mathrm{in\_lanelet}(\boldsymbol{x}, l_i)$ that specifies whether the ego vehicle is occupying a particular lane $L_j$, determined by checking its associated lanelets (cf.~Fig.~\ref{fig:cr_scenario}), 
with $\mathrm{in\_lanelet}(\boldsymbol{x}, l_i)\leftrightarrow\mathrm{occ}(\boldsymbol{x})\cap \mathrm{occ}(l_i)\not=\emptyset$ \cite[Sec.~V]{liu2023specification}. The action \textsc{follow-lane} is interpreted as the ego vehicle staying in the currently occupied lane $L_{\mathrm{cur}}$. We concretize the lane change actions \textsc{left-lane} and \textsc{right-lane} as the ego vehicle eventually remaining in the corresponding adjacent lane $L_{\mathrm{adj\_l}}$ or $L_{\mathrm{adj\_r}}$ within the decision-making horizon; thus, $\mathbf{F}\mathbf{G}$ is used as the outermost temporal operator.
By translating the LLM action pairs into LTL$_\text{f}$ formulas, we obtain a sequence of specifications $\varphi_1, \varphi_2, \ldots,\varphi_{\kappa}$ through the conjunction of the longitudinal and lateral actions. 

\vspace{2mm}
\noindent \textbf{Running Example:} The LTL$_\text{f}$ formula corresponding to the action pair \textsc{accelerate} and \textsc{follow-lane} is formalized as
\begin{equation}\label{eq:acc_follow}
	\varphi_{\text{acc\_foll}} = \mathbf{G}(a > a_{\mathrm{lim}}) \land  \mathbf{G}\big(\mathrm{in\_lane}(\boldsymbol{x}, L_{\mathrm{cur}})\big).
\end{equation}
\subsection{Formalized Traffic Rules} \label{subsec:rules}

Traffic rules cover both general regulations (indexed by R\_G) \cite[Sec.~III-A]{Maierhofer2020a} and rules for specific use cases or road types, such as interstate  \cite[Sec.~III-B]{Maierhofer2020a} and intersection  \cite{Maierhofer2022a}.
Because embedding traffic rules into prompts may lead to ambiguous interpretations and relies on the LLM to reason implicitly, we explicitly incorporate them into the verification process to ensure legal safety.
To achieve this, each action specification  $\varphi_1, \varphi_2, \ldots,\varphi_{\kappa}$ is individually augmented by conjoining it with these formalized rules expressed in LTL$_\text{f}$~\cite{Esterle2020}, which are themselves conjuncted.  When these rules are formalized in other temporal logics, such as MTL \cite{Maierhofer2020a, Maierhofer2022a}, they can be rewritten in LTL$_{\text{f}}$ using conversion tools, e.g., Spot \cite{duret2016spot}.

\vspace{2mm}
\noindent \textbf{Running Example:} Let us extend the running example by additionally considering the German Road Traffic Regulation (StVO), where we select the safe distance rule R\_G1 as an example~\cite[Sec.~III-A]{Maierhofer2020a} (translated from German): 
\begin{displayquote}
	\textit{The distance to a vehicle ahead must generally be large enough that one can stop safely even if that vehicle brakes suddenly. [...] (§~4(1) StVO).}
\end{displayquote}
\noindent Given the reaction time $t_d$ of the ego vehicle, the \textit{large enough} safe distance is defined as \cite[Sec.~IV-C]{Maierhofer2020a}:
\begin{equation}\label{eq:safe_dis}
	d_{\text{safe}}(v, {v}_{\text{obs}}) = \frac{v_{\text{obs}}^2}{-2|a_{\text{obs},\text{min}}|} - \frac{v^2}{-2|a_{\text{min}}|} + vt_d,
\end{equation}
where ${v}_{\text{obs}}$ denotes the velocity of the obstacle $\mathrm{obs}$, and $a_{\text{min}}$ and $a_{\text{obs},\text{min}}$ represent the minimum accelerations of the ego and leading vehicles, respectively, with $a_{\text{obs},\text{min}} < a_{\text{min}} < 0$. The concretized safe distance rule can be formalized in LTL$_{\text{f}}$ as \cite[(2)]{nomoretraffic}:\footnote{For simplicity, the case without a prior cut-in is omitted. Details of the rule formalization process can be found in~\cite[Sec.~IV]{nomoretraffic}.}
\begin{equation}\label{eq:rg1}
	\begin{aligned}
			\varphi_{\text{R\_G1}} = \mathbf{G}\big(\mathrm{precedes}( &\boldsymbol{x}_{\text{obs}}, \boldsymbol{x})\rightarrow \\ & \mathrm{keeps\_safe\_distance(\boldsymbol{x}, \boldsymbol{x}_{\text{obs}})}\big),
	\end{aligned}
\end{equation}
where the meaning of the predicate $\mathrm{precedes}(\boldsymbol{x}_{\text{obs}}, \boldsymbol{x})$ is self-explanatory, and $\mathrm{keeps\_safe\_distance}(\boldsymbol{x}, \boldsymbol{x}_{\text{obs}})$ holds if and only if the relative longitudinal distance  $\Delta s$ between the ego vehicle and the preceding vehicle $\mathrm{obs}$ exceeds the threshold $d_{\text{safe}}$ (cf.~(\ref{eq:safe_dis})):
\begin{equation*}
	\mathrm{keeps\_safe\_distance(\boldsymbol{x}, \boldsymbol{x}_{\text{obs}})} \leftrightarrow \Delta s > d_{\text{safe}}(v, v_{\text{obs}}).
\end{equation*}

\subsection{Safety Verification}\label{subsec:safety_v}

\begin{figure}[t!]
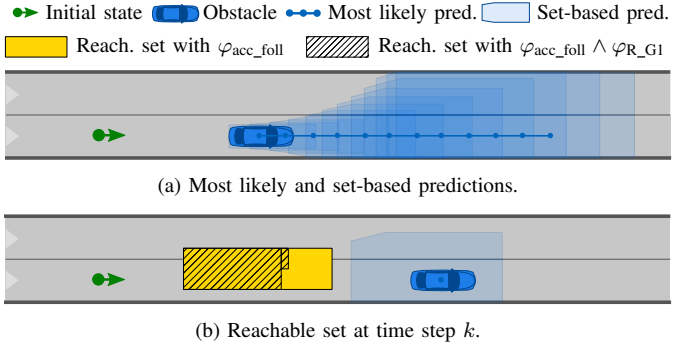

	\vspace{1.5mm}
	\centering\footnotesize
	\begin{subfigure}{1\columnwidth}
		\centering
		\includesvg[width=\linewidth]{figures/set_pre.svg}
		\caption{Most likely and set-based predictions.}
		\label{fig:set_pre}
	\end{subfigure}\vspace{2mm}
	\hfill
	\begin{subfigure}{1\columnwidth}
		\centering
		\includesvg[width=\linewidth]{figures/reach_pre.svg}
		\caption{Reachable set at time step $k$.}
		\label{fig:reach}
	\end{subfigure}
	\caption{Prediction of other traffic participants and reachable set of the ego vehicle. The specifications $\varphi_{\text{acc\_foll}}$ and $\varphi_{\text{R\_G1}}$ are given in (\ref{eq:acc_follow}) and (\ref{eq:rg1}), respectively.} 
\vspace{-3mm}
\label{fig:safety_ver}
\end{figure}
 
 The online safety verification relies on dynamic models that \textcolor{red}{predict the future behaviors} of both the ego vehicle and other traffic participants, \textcolor{red}{capturing legally feasible future evolutions of the scenario while accounting for measurement and modeling uncertainties \cite{althoff2014online}}. For the behavior of other vehicles, we use their most likely trajectories \cite{lefevre2014survey} to prune traffic-rule-violating solution spaces \cite{liu2023specification, lercher2024specification}, while set-based predictions define the forbidden states $\mathcal{X}_k^{\mathtt{F}}$ to ensure safety during decision-making (cf.~(\ref{eq:reach})) \cite{Koschi2020}. 
Examples of both types of predictions are illustrated in Fig.~\ref{fig:set_pre}. These predictions serve complementary purposes: the most likely trajectories primarily capture interactions with other traffic participants,  while the set-based predictions are required to ensure safety.
\textcolor{red}{However, other prediction strategies that provide either multiple trajectory hypotheses or safety-guaranteeing occupancy representations, i.e., conservative regions that overapproximate the possible future states of other agents, can also be integrated into our approach \cite{lefevre2014survey}.}
With this setup, we perform reachability analysis combined with model checking to compute the reachable sets $\mathcal{R}_k(\varphi)$ for $k\in[0, h]$ that satisfy the joint action and traffic-rule specification $\varphi$ \cite{liu2023specification, lercher2024specification}. 
We iterate through the specification sequence until we identify an action pair that is verified as safe, which is then returned. 

To ensure safety over an infinite time horizon, we compute fail-safe solutions in parallel that can always safely transfer the ego vehicle to an invariably safe state \cite{pek2018efficient}, i.e., a state that remains safe for an infinite time horizon. \textcolor{red}{This guarantees collision-free operation for all behaviors of other traffic participants that are consistent with the assumed traffic rules and modeling assumptions \cite[Sec.~X]{nomoretraffic}.}
\textcolor{red}{An example of such an invariably safe state is engaging a provably correct adaptive cruise controller \cite{althoff2020provably} at the end of the planning horizon on highways. We refer to the inductive proof in \cite[Sec.~III]{pek2020fail} showing that the verification pipeline guarantees that the ego vehicle remains within a safe state at all times and that a fail-safe solution is always available.}

\vspace{2mm}
\noindent \textbf{Running Example:} As shown in Fig.~\ref{fig:reach}, the reachable set at time step $k$ with respect to $\varphi_{\text{acc\_foll}}$ (cf.~(\ref{eq:acc_follow})) considers the set-based prediction of the preceding vehicle as the forbidden states, with colliding regions removed. By further considering 
$\varphi_{\text{R\_G1}}$ along with the most likely prediction of the preceding vehicle (cf.~(\ref{eq:rg1})), reachable states that violate the safe distance rule are additionally pruned.



\section{Evaluation}\label{sec:eva}
In this section, we evaluate the effectiveness of our proposed framework under the settings described in Sec.~\ref{subsec:general}. We begin with a case study in Sec.~\ref{subsec:case}, followed by several ablation studies in Sec.~\ref{subsec:abl} to validate the design choices of the \textsc{SanDRA} framework. Finally, we assess its performance in a closed-loop decision-making setting in Sec.~\ref{subsec:closed}.
\renewcommand{\arraystretch}{1.15}
\begin{table}[!t]\centering\footnotesize
	\vspace{1.5mm}
	\caption{General parameters for the evaluations.} 
	\begin{tabular}{@{}ll@{}} \toprule
		\textbf{Description} & \textbf{Notation and value}\\ \midrule
		Threshold  & $a_{\mathrm{lim}} = 0.2\mathrm{m/s^2}$;  $v_{\mathrm{err}} = 0.1\mathrm{m/s}$  \\ 
		Reachability analysis & \makecell[l]{longitudinal acceleration: $[-6, 6]\mathrm{m/s^2}$;\\
			lateral acceleration: $[-4, 4]\mathrm{m/s^2}$;\\ maximum velocity: $30\mathrm{m/s}$}\\
		Set-based prediction & maximum acceleration: $12\mathrm{m/s^2}$\\
		Safe distance $d_{\text{safe}}$  & \makecell[l]{$t_d=0.4\unit{s}$;  $a_{\text{min}}=-6\mathrm{m/s^2}$; $a_{\text{obs},\text{min}}=-12\mathrm{m/s^2}$}  \\ \midrule
		\multicolumn{2}{l}{\noindent\bf{Dataset evaluation}} \\
		Temporal setting &  $\Delta t = 0.04\unit{s}$\\\vspace{1mm}
		 \makecell[l]{\textbf{Longitudinal}: $0.0\%$ \textsc{stop}, $55.8\%$ \textsc{keep},\\
    $21.4\%$ \textsc{accelerate}, $22.8\%$ \textsc{decelerate}\\
    \textbf{Lateral}: $98.7\%$ \textsc{follow-lane},\\
    $0.9\%$ \textsc{left-lane}, $0.4\%$ \textsc{right-lane}
}\\
		 \makecell[l]{\textbf{Longitudinal}: $0.0\%$ \textsc{stop}, $62.1\%$ \textsc{keep},\\
    $24.3\%$ \textsc{accelerate}, $13.6\%$ \textsc{decelerate}\\
    \textbf{Lateral}: $98.4\%$ \textsc{follow-lane},\\
    $1.0\%$ \textsc{left-lane}, $0.6\%$ \textsc{right-lane}
}\\
		\midrule
		\multicolumn{2}{l}{\noindent\bf{Highway-env simulation}} \\
		\multicolumn{2}{l}{\hspace{-0.7em}Action mapping (highway-env vs. our approach)} \\
		\enspace LANE\_LEFT & \makecell[l]{\textsc{left-lane}, \textsc{keep}/\textsc{decelerate}/\textsc{accelerate}} \\
		\enspace IDLE & \textsc{follow-lane}, \textsc{keep}\\
		\enspace LANE\_RIGHT & \makecell[l]{\textsc{right-lane}, \textsc{keep}/\textsc{decelerate}/\textsc{accelerate}} \\
		\enspace FASTER & \textsc{follow-lane}, \textsc{accelerate}\\
		\enspace SLOWER & \textsc{follow-lane}, \textsc{decelerate}\\
		Hyperparameter & $\kappa = 3$; number of historical actions: $5$ \\
		Frequency setting & simulation: $15\unit{Hz}$; policy: $5\unit{Hz}$\\
		Temporal setting & \makecell[l]{scenario duration: $30$ steps; $\Delta t = 0.2\unit{s}$; \\ $h=15$  (most likely), $h=8$ (set-based)} \\
		\bottomrule
	\end{tabular}\vspace{-1mm}
	\label{tab:params}
\end{table}
\subsection{General Settings}\label{subsec:general}
We evaluated \textsc{SanDRA} using the open-source CommonRoad framework \cite{Althoff2017a}. In our implementation, we used the CommonRoad-Reach toolbox \cite{reach} for reachability analysis and the Spot tool \cite{duret2016spot} for LTL$_{\text{f}}$ model checking. \textcolor{red}{In the reachability analysis, we employed a point-mass model that overapproximated the vehicle dynamics, conservatively subsuming all feasible behaviors by using bounded accelerations as inputs to compute the reachable sets.} Additionally, we used time-to-collision as the criticality measure, computed with the CommonRoad-CriMe toolbox \cite{lin2023commonroad}.  In closed-loop evaluations, the widely-used highway-env \cite{highway-env} simulation environment was employed to assess the safety of driving actions, where the most likely trajectories were predicted under a constant-velocity assumption. The set-based predictions of the other vehicles were computed based on \cite{Koschi2020}. We used GPT-4o \cite{hurst2024gpt} as the cloud-based LLM and Qwen3 \cite{qwen3} as the local LLM.  The experiments were conducted on a desktop equipped with an AMD Ryzen 7 9800X 8-core processor and an NVIDIA RTX 5090 GPU. The parameters used in the numerical experiments are shown in Tab.~\ref{tab:params}; all remaining parameters were as defined in the original paper.


\begin{figure}[t!]
	\centering\footnotesize
	\begin{subfigure}{1\columnwidth}
		\centering
		\includesvg[width=\linewidth]{figures/initial.svg}
		\caption{Initial scenario.}
		\label{fig:subfig_initial}
	\end{subfigure}\vspace{1mm}
	\hfill
	\begin{subfigure}{1\columnwidth}
		\centering
		\includesvg[width=\linewidth]{figures/reach.svg}
		\caption{Reachable sets for the actions \textsc{accelerate} and \textsc{right-lane}.}
		\label{fig:acc_fail}
	\end{subfigure}\vspace{1mm}
	\begin{subfigure}{1\columnwidth}
		\centering
		\includesvg[width=\linewidth]{figures/final_plan.svg}
		\caption{Reachable sets for the actions \textsc{decelerate} and \textsc{follow-lane} as well as the planned trajectory.}
		\label{fig:plan}
	\end{subfigure}\vspace{1mm}
	\begin{subfigure}{1\columnwidth}
		\centering
		\includesvg[width=\linewidth]{figures/sonia.svg}
		\caption{Reachable sets for the actions \textsc{decelerate} and \textsc{follow-lane} with the set-based prediction of the other obstacles.}
		\label{fig:sonia}
	\end{subfigure}\vspace{1mm}
	\begin{subfigure}{1\columnwidth}
		\centering
		\includesvg[width=\linewidth]{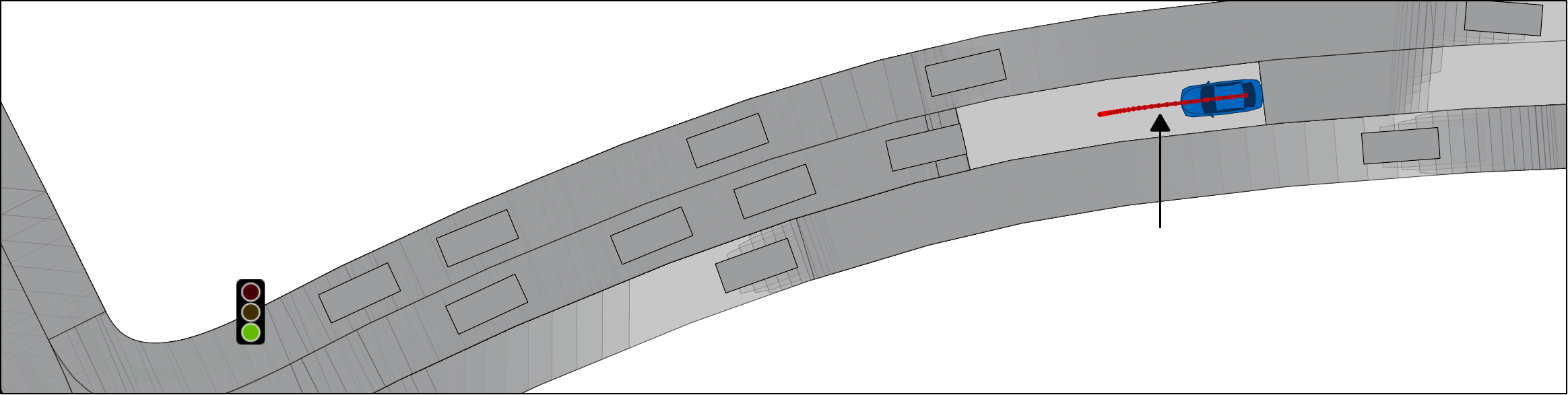}
		\caption{Fail-safe trajectory when no verified action is available, with the velocity profile decelerating to a standstill.}
		\label{fig:fail-safe}
	\end{subfigure}
	\caption{CommonRoad scenarios\protect\footnotemark\ in which the ego vehicle approaches a signalized intersection. (\ref{fig:acc_fail}) and (\ref{fig:plan}) use the most likely predictions of other obstacles, while (\ref{fig:sonia}) employs set-based predictions. When formalized traffic rules are incorporated in (\ref{fig:fail-safe}), no safe action is found and a fail-safe mode is activated. The trajectory of the ego vehicle is referenced from the rear axle, and, in the 3D view, the upward axis denotes time steps.}\vspace{-4mm}
	\label{fig:commonroad_sce}
\end{figure}
\subsubsection{Scenarios}

	For open-loop batch evaluation, we used \textcolor{red}{highway} scenarios from the highD dataset \cite{highd} to finetune LLMs and \textcolor{red}{urban} scenarios from the MONA dataset \cite{gressenbuch2022mona} to evaluate the performance of our approach. Both datasets contained recordings of naturalistic vehicle trajectories, where the trajectories of other obstacles were treated as their most likely predictions. \textcolor{red}{For each scenario, a vehicle was randomly selected as the ego vehicle, a decision-making problem was formulated based on its initial state, and the vehicle was subsequently removed from the scenario.} The trajectory of the ego vehicle was labeled with high-level actions based on the LTL$_{\text{f}}$ formulas in Tab.~\ref{tab:action}, \textcolor{red}{whose distribution was summarized in Tab.~\ref{tab:params}.} 
\subsubsection{Traffic Rules}\label{subsec:rule_intro}\footnotetext{CommonRoad-ID: DEU\_Goeppingen-37\_1\_T-4}
We adopted the German traffic rules from \cite[Sec.~III-A]{Maierhofer2020a}, which were formalized in MTL and rewritten in LTL$_{\text{f}}$. In addition to the safe distance rule R\_G1 introduced in the running example (cf.~(\ref{eq:rg1})), the following general traffic rules were considered (see Fig.~\ref{fig:prompt}):
\begin{itemize}
\item R\_G2: \enquote{\textit{The ego vehicle is not allowed to brake abruptly without reason.}}, which was formalized as \cite[(11)]{YuanfeiLin2022a}:
\begin{equation}\label{eq:rg2}
	\begin{aligned}
		\varphi_{\text{R\_G2}} = \mathbf{G}\big(\mathrm{brakes\_}&\mathrm{abruptly}(\boldsymbol{x})\rightarrow \\ & \mathrm{braking\_justification}(\boldsymbol{x})\big),
	\end{aligned}
\end{equation}
where we define $\mathrm{braking\_justification}(\boldsymbol{x})$ as the case in which the ego vehicle executes a fail-safe action.

\item R\_G3: \enquote{\textit{The ego vehicle must not exceed the speed limit.\footnote{The speed limit value was automatically obtained from the environment, e.g., from traffic signs, and integrated into the rule description as \enquote{the maximum allowed speed is \highlight{speed limit}}.}}}, which was formalized as \cite[Tab.~I]{Maierhofer2020a}:

\begin{equation}\label{eq:rg3}
	\begin{aligned}
		\varphi_{\text{R\_G3}} = \mathbf{G}\big(&\mathrm{keeps\_lane\_speed\_limit}(\boldsymbol{x})\land \\ & \mathrm{keeps\_fov\_speed\_limit}(\boldsymbol{x})\land \\ &
		\mathrm{keeps\_type\_speed\_limit}(\boldsymbol{x})\land \\ &
		\mathrm{keeps\_braking\_speed\_limit}(\boldsymbol{x})\big).
	\end{aligned}
\end{equation}
\end{itemize}

\subsection{Case Study}\label{subsec:case}

Consider a German urban scenario (cf.~Fig.~\ref{fig:subfig_initial}) where the ego vehicle was heading toward a signalized intersection with a velocity of $7.7\unit{m/s}$ and acceleration of $-2.1\unit{m/s^2}$. Given the lane structure -- with a right adjacent lane to the ego vehicle -- all longitudinal actions were feasible, while the available lateral actions were limited to \textsc{follow-lane} and \textsc{right-lane}. The user command was set to \enquote{\it drive faster}, with the number of ranked action pairs limited to $\kappa=3$, a decision-making horizon of $h=25$, and a time increment of $\Delta t = 0.1\unit{s}$. After automatically describing the environment and providing the prompt to GPT-4o, we obtained a ranked sequence of three longitudinal-lateral action pairs:

\begin{enumerate}
	\item \textsc{accelerate}, \textsc{right-lane},
	\item \textsc{keep}, \textsc{right-lane}, and
	\item \textsc{decelerate}, \textsc{follow-lane}.
\end{enumerate}

We verified the safety of the first action pair by computing the reachable sets that satisfy the corresponding action formula in LTL$_\text{f}$, as shown in Fig.~\ref{fig:acc_fail}. The reachable set at time step~$23$ was notably small and became empty at time step~$24$ due to complete overlap with the predicted occupancy of the preceding vehicle. As a result, the first action pair was verified to be unsafe. The second-best result, \textsc{keep} and \textsc{right-lane}, likewise resulted in empty reachable sets due to collisions. Afterwards, we computed the reachable sets satisfying the last action pair \textsc{decelerate} and \textsc{follow-lane}, as shown in Fig.~\ref{fig:plan}, and used a sampling-based planner~\cite{wursching2021sampling} to find the optimal trajectory within these sets. By sampling only within the obtained reachable sets, the trajectory was found faster than sampling the entire solution space. 

Fig.~\ref{fig:sonia} shows that the last action pair was provably safe within the decision-making horizon, provided the reachable sets remained non-empty under the set-based prediction of other obstacles. However, the action pair was not legally safe, as the reachable sets became empty when the action formula was conjoined with $\varphi_{\text{R\_G1}}\land\varphi_{\text{R\_G2}}\land\varphi_{\text{R\_G3}}$. This occurred because the initial velocity of the ego vehicle was too high for it to brake smoothly and avoid a collision if the preceding vehicle braked suddenly; that is, the safe distance rule $\varphi_{\text{R\_G1}}$ could not be satisfied without violating the unnecessary braking rule $\varphi_{\text{R\_G2}}$. In such a situation, a fail-safe solution was required to guarantee safety over an infinite time horizon, bringing the ego vehicle to a standstill, as shown in Fig.~\ref{fig:fail-safe}.

\subsection{Ablation Study}\label{subsec:abl}
\begin{figure}[t!]
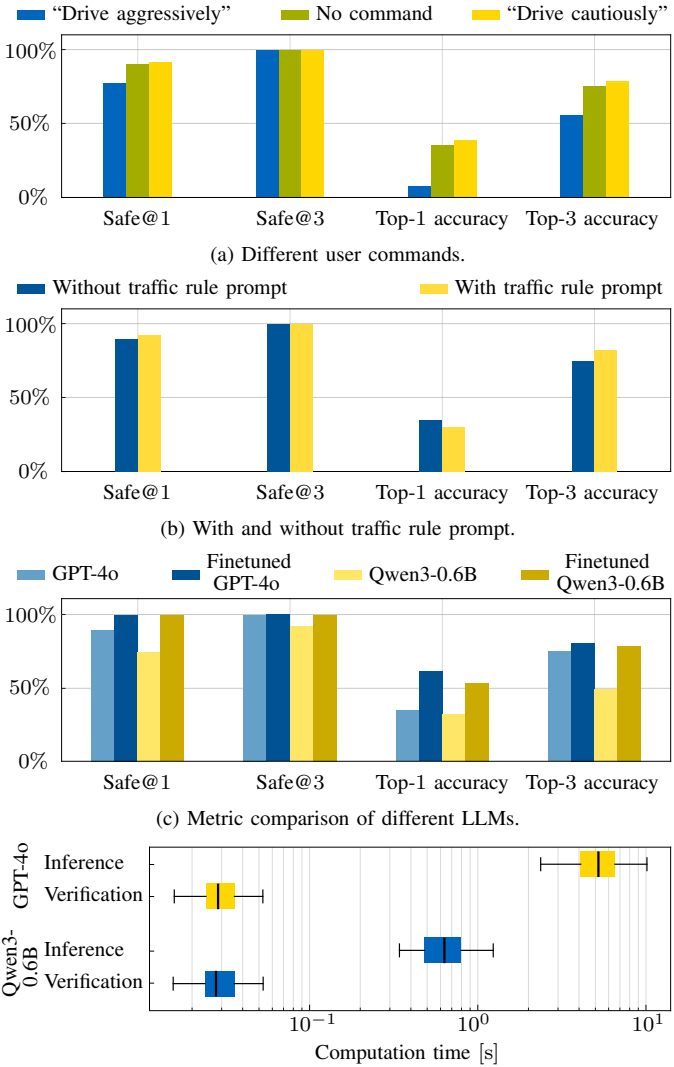

	\centering\footnotesize
		\begin{subfigure}{1\columnwidth}
		\vspace{1.5mm}
		\centering
		\includesvg[width=\linewidth]{figures/horizon-comp}
		\caption{\textcolor{red}{Different decision-making horizons.}}
		\label{fig:horizon}
	\end{subfigure}
	\begin{subfigure}{1\columnwidth}
		\vspace{2mm}
		\centering
		\includesvg[width=\linewidth]{figures/command-comp}
		\caption{Different user commands.}
		\label{fig:command}
	\end{subfigure}
	\hfill
		\begin{subfigure}{1\columnwidth}
		\vspace{2mm}
		\centering
		\includesvg[width=\linewidth]{figures/rule_prompt}
		\caption{With and without traffic rule prompt.}
		\label{fig:rule_prompt}
	\end{subfigure}
	\hfill
	\begin{subfigure}{1\columnwidth}
		\centering
		\vspace{2mm}
		\includesvg[width=\linewidth]{figures/llm-comp}
		\caption{Metric comparison of different LLMs.}
		\label{fig:llms}
	\end{subfigure}\hfill
	\begin{subfigure}{1\columnwidth}
		\centering
		\vspace{2mm}
		\includesvg[width=\linewidth]{figures/runtime-comp.svg}
		\caption{\textcolor{red}{Runtime comparison of the approaches shown in (\ref{fig:llms}) and of the exhaustive verification of all action combinations without LLM assistance. For improved clarity, outliers and runs exceeding $30\unit{s}$ are excluded.}}
		\label{fig:runtime}
	\end{subfigure}
	\caption{Ablation studies on the effectiveness of the main components of \textsc{SanDRA}.}\vspace{-2mm}
	\label{fig:histogram}
\end{figure}

We conducted ablation studies on the main components of the \textsc{SanDRA} framework, evaluating their effectiveness using two metrics in an open-loop setting \textcolor{red}{on the $800$ MONA test scenarios (cf. Tab.~\ref{tab:params})}:
\begin{itemize}
	\item \textbf{Safe@$\boldsymbol{\kappa}$}: the proportion of scenarios where at least one of the returned $\kappa$ action pairs was verified as safe by \textsc{SanDRA}, using the recorded trajectories of other obstacles; and
	\item \textbf{Top-$\boldsymbol{\kappa}$ accuracy}: the proportion of scenarios where the dataset label appeared among the top $\kappa$ action pairs.
\end{itemize}
 Please note that the final state recorded in the dataset was withheld from the LLM to avoid revealing the ground-truth decision outcome. Moreover, we did not include the formalized traffic rules in the reachability analysis, as it was shown that many human drivers violated rules in the dataset \cite{Maierhofer2020a, Maierhofer2022a}, making it unsuitable as an upper-limit performance indicator.

\subsubsection{Decision-Making Horizon $h$} \textcolor{red}{In Fig.~\ref{fig:horizon}, we present the evaluation results under different decision-making horizons $h$ ($10$, $25$, and $50$), without user commands and using GPT-4o as the LLM. Although longer horizons imposed stricter safety requirements, the safe@$\kappa$ results did not decrease with increasing $h$. This was because shorter horizons often prevented lane-change maneuvers from completing successfully, and in our test scenarios, the stricter requirements of longer horizons were not pronounced enough to clearly affect the trend. In contrast, the matching metric top-$\kappa$ remained largely unchanged as the horizon increased, since it was not affected by $h$. In the following ablation studies, we used $h=25$ as the default decision-making horizon to better illustrate the differences between the various setups.}
\subsubsection{User Commands} We included three different setups for driving styles: {{no command}, the command \textquote{\textit{driving aggressively}}, and the command \textquote{\textit{driving cautiously}}. All styles used GPT-4o as the underlying LLM. The results in Fig.~\ref{fig:command} showed that, on average, a safe action could be found with $\kappa=3$ in $99.8\%$ of cases,  where $100\%$ was achieved when using fail-safe planning. Since our focus was on the safety of driving decisions, $\kappa=3$ was recommended as it balanced the maximum number of iterations with the effectiveness of finding safe decisions. Among all command options, driving cautiously was the most effective for identifying safe actions at both $k=1$ and $k=3$. In addition, the top-$\kappa$ accuracy results indicated that the behaviors of human drivers in the MONA dataset most closely aligned with a cautious driving style, which was consistent with typical driving patterns in urban environments. 
	Because the off-the-shelf LLM a) was not pretrained on structured driving data, b) received inputs in a textual format (which human drivers typically do not use for decision-making), and c) had to generate explicit driving actions, its top-$1$ accuracy remained relatively low across all commands.
\subsubsection{Traffic Rule Prompt} We compared our prompt design with and without the narrative concretization of three general traffic rules using GPT-4o as the underlying LLM without user commands. The concretization, illustrated below, is appended to the bottom of the LLM prompt (cf.~Fig.~\ref{fig:prompt}).
	\begin{mdframed}[backgroundcolor=TUMblue3,   innerleftmargin=10pt,   
	innerrightmargin=10pt, linecolor=white,      
	linewidth=0pt]\small
	\textbf{{\textless Traffic rules\textgreater}}: Please adhere to the traffic regulations in \highlight{\{country\}}:
	
	\qquad [(R\_G1: \highlight{Safe distance to preceding vehicle rule}.)] 
	
	\qquad [(R\_G2: \highlight{Unnecessary braking rule}.)]
	
	\qquad [(R\_G3: \highlight{Maximum speed limit rule}.)] 

\end{mdframed}	
As shown in Fig.~\ref{fig:rule_prompt}, including the traffic rules in the prompt made it slightly easier to identify safe actions among the first returned action pairs. With respect to accuracy against the dataset labels, the presence or absence of the traffic rule prompt did not make any noticeable difference. Given this minor effect and the fact that not all safe actions were traffic-rule-compliant, we chose to not include the traffic rule prompt in our framework. Instead, it was necessary to incorporate formalized traffic rules into safety verification to enhance legal safety in decision-making. 
\begin{table}[t!]
	\begin{center}
		\vspace{1.5mm}
		\caption{\textcolor{red}{Evaluation of safety and matching performance by action type, with the best values in each column highlighted in bold according to the action direction.}}
		\label{tab:action_type_eval}
		\begin{tabular}{@{}lcccc@{}}
			\toprule[1.pt]
			\textbf{Action} &   \enspace\textbf{Safe@$1$} \enspace& \enspace\textbf{Safe@$3$}\enspace & \textbf{Top-$1$ Acc.} & \textbf{Top-$3$ Acc.} \\\midrule
			\textsc{Keep}       & ${90.8\%}$ & $98.8\%$   & $\boldsymbol{51.8\%}$ & $\boldsymbol{85.0\%}$ \\
			\textsc{Accelerate} & $85.5\%$ & $97.9\%$    & $11.0\%$ & $35.4\%$ \\\vspace{1mm}
			\textsc{Decelerate} & $\boldsymbol{92.0\%}$ & $\boldsymbol{99.0\%}$      & $25.0\%$ & $68.2\%$ \\
			\textsc{Follow-lane}& $\boldsymbol{90.5\%}$ & $\boldsymbol{98.8\%}$ & $\boldsymbol{40.4\%}$ & $\boldsymbol{74.9\%}$ \\
			\textsc{Left-lane}  & $83.9\%$ & $98.2\%$& $1.8\%$  & $33.9\%$ \\
			\textsc{Right-lane} & $74.3\%$ & $97.1\%$ & $0.0\%$  & $20.0\%$ \\
			
			\bottomrule
		\end{tabular}\vspace{-3mm}
	\end{center}
\end{table}
\renewcommand{\arraystretch}{1.2}
\setlength{\tabcolsep}{4.8pt} 
\begin{table*}[t!]\footnotesize
	\begin{center}
		\vspace{2mm}
		\caption{Closed-loop evaluation across three settings, where values in bold denote the best performance. Note that the traveled distance is computed only for runs that were successfully completed.}
		\label{tab:performance}
		\begin{tabular}{@{}lccccccccc@{}} 
			\toprule[1.pt]
			\multirow{2}{*}{\textbf{Method}} & \multirow{2}{*}{\textbf{Setting}} & \multirow{2}{*}{\textbf{Success} \textbf{rate}} & \multirow{2}{*}{\textbf{Collision-free}  \textbf{steps}} & 	\multicolumn{3}{c}{\textbf{Rule-compliant steps}} & \multirow{2}{*}{\textbf{Success steps}} & \multirow{2}{*}{\textbf{Traveled} \textbf{distance}} & \multirow{2}{*}{\textbf{Fail-safe} \textbf{rate}}\\
			& & & & $\varphi_{\text{R\_G1}}$ & $\varphi_{\text{R\_G2}}$ & $\varphi_{\text{R\_G3}}$ & & & \\\midrule
			\multirow{3}{*}{DiLu \cite{wen2023dilu}} & \small{\textcircled{\scriptsize{1}}} & $93.3\%$ & $27.7$ & $15.7$ & $24.4$ & $27.7$ & $12.3$ & $\boldsymbol{116.5\unit{m}}$  & - \\
			& \small{\textcircled{\scriptsize{2}}} & $53.3\%$ & $18.8$ & $12.1$ & $14.7$ & $18.8$ & $8.0$ & $\boldsymbol{94.8\unit{m}}$ & - \\\vspace{1mm}
			&  \small{\textcircled{\scriptsize{3}}} & $76.7\%$ & $23.8$ & $13.8$ & $19.6$ & $23.8$ & $9.6$ & $\boldsymbol{102.4\unit{m}}$ & - \\
			\multirow{3}{*}{\makecell[l]{\textsc{SanDRA} +  \\ \enspace most likely pred. + \\ \enspace w/o traffic rules in verif.}}  & \small{\textcircled{\scriptsize{1}}} & $83.3\%$ & ${28.5}$ & $23.3$ & $21.7$ & $28.5$ & $15.3$ &  $114.4\unit{m}$ & $6.5\%$ \\
			& \small{\textcircled{\scriptsize{2}}} & $73.3\%$ & $25.6$ & $17.5$ & $19.5$ & $25.6$ & $10.9$ &  ${93.7\unit{m}}$ & $\boldsymbol{6.8\%}$ \\\vspace{1mm}
			& \small{\textcircled{\scriptsize{3}}} & $83.3\%$ & $25.8$ & $16.9$ & $20.6$ & $ 25.8$ & $11.1$&  $98.7\unit{m}$ & $\boldsymbol{7.3\%}$ \\
			\multirow{3}{*}{\makecell[l]{\textsc{SanDRA} +  \\ \enspace most likely pred. + \\ \enspace w/ traffic rules in verif.}}  & \small{\textcircled{\scriptsize{1}}} & $\boldsymbol{100.0\%}$ & $\boldsymbol{30.0}$ & $\boldsymbol{28.5}$ & $\boldsymbol{26.8}$ & $\boldsymbol{30.0}$ & $20.0$ &  $84.4\unit{m}$ & $25.2\%$ \\
			& \small{\textcircled{\scriptsize{2}}} & $96.7\%$ & $29.4$ & $27.2$ &  $\boldsymbol{26.3}$ & $29.4$ & $19.9$ &  ${68.4\unit{m}}$ & $25.0\%$ \\\vspace{1mm}
			& \small{\textcircled{\scriptsize{3}}} & $\boldsymbol{100.0\%}$ & $\boldsymbol{30.0}$ & $27.5$  & $\boldsymbol{27.2}$ & $\boldsymbol{30.0}$ & $20.3$ &  $72.5\unit{m}$ & $26.1\%$ \\
			\multirow{3}{*}{\makecell[l]{\textsc{SanDRA} +\\\enspace set-based pred. + \\ \enspace w/o traffic rules in verif.}} & \small{\textcircled{\scriptsize{1}}} & $\boldsymbol{100.0\%}$ &  $\boldsymbol{30.0}$ & $25.0$ & $16.2$ & $\boldsymbol{30.0}$ & $16.5$ &  $113.6\unit{m}$ & $\boldsymbol{3.8\%}$ \\
			&  \small{\textcircled{\scriptsize{2}}} & $\boldsymbol{100.0\%}$ & $\boldsymbol{30.0}$ & $21.4$ & $14.3$ & $\boldsymbol{30.0}$ & $14.1$ & $88.6\unit{m}$ & ${7.4\%}$ \\\vspace{1mm}
			&  \small{\textcircled{\scriptsize{3}}} & $\boldsymbol{100.0\%}$ & $\boldsymbol{30.0}$ & $19.6$ & $13.7$ & $\boldsymbol{30.0}$ & $13.1$ &  $86.4\unit{m}$ & ${7.9\%}$ \\	
			\multirow{3}{*}{\makecell[l]{\textsc{SanDRA} +\\\enspace set-based pred. + \\ \enspace w/ traffic rules in verif.}} & \small{\textcircled{\scriptsize{1}}} & $\boldsymbol{100.0\%}$ &  $\boldsymbol{30.0}$ & $28.4$ & $25.3$ & $\boldsymbol{30.0}$ & $\boldsymbol{20.8}$ &  $85.6\unit{m}$ & ${25.9\%}$ \\
			&  \small{\textcircled{\scriptsize{2}}} & $\boldsymbol{100.0\%}$ & $\boldsymbol{30.0}$ & $\boldsymbol{27.7}$ & $23.9$ & $\boldsymbol{30.0}$ & $\boldsymbol{20.9}$ & $70.1\unit{m}$ & ${26.9\%}$ \\
			&  \small{\textcircled{\scriptsize{3}}} & $\boldsymbol{100.0\%}$ & $\boldsymbol{30.0}$ & $\boldsymbol{27.6}$ & $23.3$ & $\boldsymbol{30.0}$  & $\boldsymbol{20.2}$ &  $72.2\unit{m}$ & ${25.8\%}$ \\	
			\bottomrule
		\end{tabular}\vspace{-2mm}
	\end{center}
\end{table*}
\subsubsection{LLMs} 
\textcolor{red}{We conducted supervised finetuning of both GPT-4o and Qwen3-8B on the highD dataset.} Each training example consisted of a system prompt and a user prompt describing a driving scenario (cf.~Fig.~\ref{fig:prompt}), deliberately excluding explicit user commands.
The output was a JSON-formatted action ranking, with the labeled action pair ranked highest and the remaining positions filled with randomly ordered feasible action pairs. GPT-4o was finetuned via the OpenAI API, while low-rank adaptation \cite{hu2022lora} was applied to the locally hosted Qwen3 model. \textcolor{red}{Both models used instruction finetuning on the collected training samples, with the objective of predicting the labeled action rankings from the scenario prompt, using cross-entropy loss over the token sequence.} As shown in Fig.~\ref{fig:llms}, both finetuned models achieved high safe@$1$ and safe@$3$ scores exceeding $99\%$, along with a significant increase in top-$1$ accuracy compared to their pretrained counterparts.  Additionally, Qwen3 achieved comparable top-$\kappa$ accuracy to GPT-4o despite having a much smaller model size. The runtime performance of the GPT-4o and Qwen3-0.8B models with $\kappa=3$, including both their original and finetuned versions, was shown in Fig.~\ref{fig:runtime}. Qwen3-0.8B achieved an average computation time of $0.67\unit{s}$, which was considerably shorter than that of GPT-4o at $5.56\unit{s}$, and was suitable for local deployment with low latency. For both models, safety verification consumed only a negligible portion of the overall decision-making process, with an average runtime of $0.033\unit{s}$ and a standard deviation of $0.018\unit{s}$.
\textcolor{red}{We additionally reported the runtime of randomly enumerating all $12$ action combinations without LLM assistance. This exhaustive computation was more expensive than verifying the top-$3$ action pairs suggested by the LLM, yet still faster than the LLM inference itself, indicating potential room for improvement in online LLM acceleration.}
\subsubsection{Action Type} \textcolor{red}{We categorized the safe@${\kappa}$ and top-$\kappa$ accuracy metrics from the above ablation studies by action type (cf. Tab.~\ref{tab:action}) in Tab.~\ref{tab:action_type_eval} to illustrate how different actions impact safety and the accuracy of action predictions. The results indicated that the $\textsc{decelerate}$ and $\textsc{follow-lane}$ actions achieved the highest safe@${\kappa}$ score in the longitudinal and lateral directions, respectively, which was aligned with the safe distance rule $\varphi_{\text{R\_G1}}$, as deceleration maintained longitudinal spacing and lane following protected lateral safety. Since the majority of driving behaviors in the training dataset involved maintaining velocity and following the lane (cf. Tab.~\ref{tab:params}), the $\textsc{keep}$ and $\textsc{follow-lane}$ actions achieved the highest top-$\kappa$ accuracy among all actions.}
\subsection{Closed-Loop Evaluation} \label{subsec:closed}
In the highway-env simulation, our longitudinal and lateral action pairs were mapped to the embedded discrete meta-actions, as listed in Tab.~\ref{tab:params}.
Note that the longitudinal action \textsc{stop} was excluded, as it was not applicable in the given highway driving context. 
Due to the absence of explicit emergency maneuvers in the simulation environment and the modeling of other vehicles using the intelligent driver model \cite{highway-env}, the meta-action {SLOWER} was adapted as the fail-safe behavior by increasing its braking intensity.\footnote{The time constant for the speed response was set to $0.6\unit{s}$, and the speed increment step was set to $15\unit{m/s}$.}
Furthermore, we utilized GPT-4o as the LLM, incorporating the user command \enquote{\textit{do not change lanes too often}} alongside the traffic rule prompt encoding the general rules listed in Sec.~\ref{subsec:rule_intro}. 
\subsubsection{Evaluation Settings}
We compared our approach to the state-of-the-art framework DiLu \cite{wen2023dilu}, which featured a decision-making agent equipped with memory, reasoning, and reflection capabilities, as well as a memory module for retaining experience. Note that we disabled memory collection during evaluation to maintain a fair comparison. Following \cite{wen2023dilu}, we adapted six metrics to evaluate the performance of safe decision-making in a closed-loop setting:
\begin{itemize}
	\item \textbf{Success rate}: the percentage of scenarios where the ego vehicle completed the full duration without collisions;
	\item \textbf{Collision-free steps}: the average collision-free time steps across all scenarios;
	\item \textbf{Rule-compliant steps}: the average number of time steps across all scenarios during which the ego vehicle remained compliant with the given traffic rule;
	\item \textbf{Success steps}: the average number of time steps across all scenarios that were both collision-free and compliant with all traffic rules;
	\item \textbf{Traveled distance}: the average longitudinal distance traveled from the initial state to the end of the scenario simulation; and
	\item \textbf{Fail-safe rate}: percentage of time steps during which a fail-safe action was executed.
\end{itemize}
The evaluation was conducted under three different environmental settings:
\begin{itemize}
	\item[\textcircled{\scriptsize{1}}] highway with four lanes and a vehicle density\footnote{The larger the density, the smaller the vehicle longitudinal spacing.} of $2$;
	\item[\textcircled{\scriptsize{2}}] highway with four lanes and a vehicle density of $3$; and
	\item[\textcircled{\scriptsize{3}}] highway with five lanes and a vehicle density of $3$.
\end{itemize}
 Each setting was evaluated using $10$ different random seeds\footnote{Highway-env seeds:     5838,
 	2421,
 	7294,
 	9650,
 	4176,
 	6382,
 	8765,
 	1348,
 	4213, and	
 	2572.} and repeated three times.
\textcolor{red}{The ego vehicle updated its decision every three simulation steps, and the selected action was held constant between updates (cf. Tab.~\ref{tab:params}).}
 	
\subsubsection{Safety Evaluation}
The results in Tab.~\ref{tab:performance} showed that all variants of \textsc{SanDRA} achieved a performance comparable to or better than DiLu in terms of success rate. This held even though it did not incorporate few-shot examples from past driving experiences into the prompt. By relying solely on the zero-shot prompt, our approach not only reduced token usage but also eliminated dependency on the quality and availability of prior examples.
Regarding formal safety, relying solely on most likely predictions was insufficient, particularly in critical scenarios where the actual behavior of the other obstacles might have deviated significantly, e.g., under setting \textcircled{\scriptsize{2}}.  Although formalized traffic rules were considered and fail-safe actions were frequently triggered, \textcolor{red}{these fallback
	interventions could still fail, i.e., when the current scenario had already left the ego vehicle without any feasible safe
	action.} 
In contrast, set-based prediction allowed the ego vehicle to account for all possible behaviors of other vehicles over the decision-making horizon, thereby ensuring safety at all times. 

\subsubsection{Rule Compliance}
For traffic rule compliance, we observed that \textsc{SanDRA} substantially improved adherence to the safe distance rule $\varphi_{\text{R\_G1}}$ compared to DiLu, irrespective of the specific variant. The greatest improvement was achieved when formalized rules were incorporated into the reachability analysis. However, this also resulted in more fail-safe actions, i.e., emergency braking, which in turn reduced the traveled distance, particularly in our dense traffic settings. When combining most likely predictions with traffic rules in the verification, compliance with $\varphi_{\text{R\_G2}}$ was generally higher, whereas set-based predictions maintained larger longitudinal gaps between vehicles, often resulting in unnecessary braking. Furthermore, all runs complied with the speed limit rule $\varphi_{\text{R\_G3}}$.
Overall, integrating set-based predictions with formalized traffic rules into the safety verification yielded the highest number of success steps, i.e., the best collision-free and legally compliant performance.
 	
\begin{figure}[t!]
	\centering\footnotesize
	\vspace{2mm}
	\includesvg[width=\linewidth]{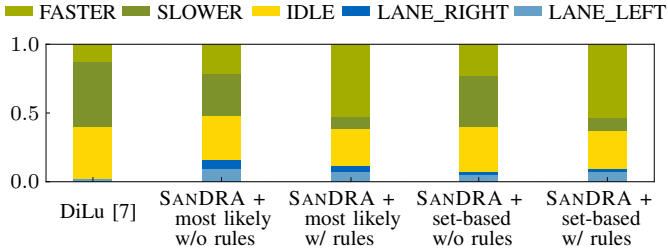}
	\caption{Action distribution of the ego vehicle in the highway-env simulation, where fail-safe actions are not taken into account.}\vspace{-2mm}
	\label{fig:distribution}
\end{figure}
\subsubsection{Conservatism}
The traveled distance reported in Tab.~\ref{tab:performance} indicated that, without integrating rules into the verification process, the set-based approach did not induce overly conservative behavior for the ego vehicle, particularly in less dense scenarios (e.g., setting {\small \textcircled{\scriptsize{1}}}). This was further supported by the low fail-safe rate and by the fact that \textsc{SanDRA} allowed more lane-change options than DiLu, regardless of the prediction model (cf. Fig.~\ref{fig:distribution}). In contrast, incorporating traffic rules in the reachability analysis increased the rate of fail-safe actions, since most highway-env scenarios were not initialized with the ego vehicle maintaining a safe distance to preceding vehicles; consequently, the first few time steps were often required for emergency braking. This also resulted in additional acceleration actions (cf. Fig.~\ref{fig:distribution}), as the ego vehicle needed to recover the distance lost during the fail-safe period. Reducing the number of fail-safe actions would have required either relaxing the traffic rules or adopting trajectory repair for rule compliance \cite{yuanfei2024}, both of which, however, were beyond the scope of this work.
\begin{figure}[t!]
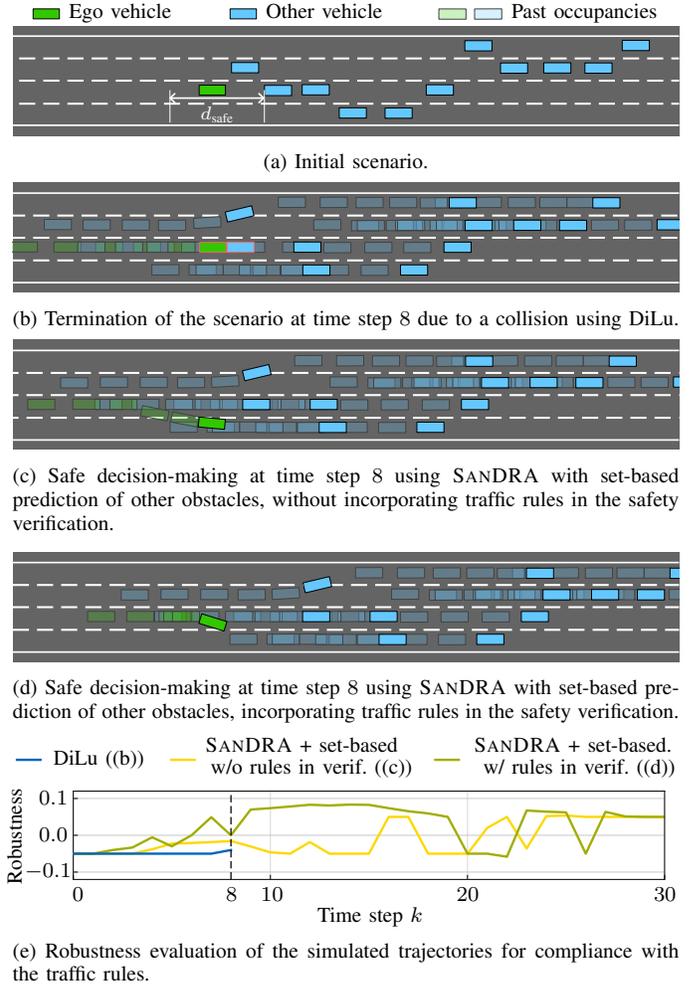

	\centering\footnotesize
	\begin{subfigure}{1\columnwidth}
		\vspace{1.5mm}
		\centering
		\includesvg[width=\linewidth]{figures/frame_0000.svg}
		\caption{Initial scenario.}
		\label{fig:highenv-ini}
	\end{subfigure}\vspace{1mm}
	\hfill
	\begin{subfigure}{1\columnwidth}
		\centering
		\includesvg[width=\linewidth]{figures/frame_0008_dilu.svg}
		\caption{Termination of the scenario at time step~$8$ due to a collision using DiLu.}
		\label{fig:highenv-dilu}
	\end{subfigure}\vspace{1mm}
	\begin{subfigure}{1\columnwidth}
		\centering
		\includesvg[width=\linewidth]{figures/frame_0008_safe.svg}
		\caption{Safe decision-making at time step~$8$ using \textsc{SanDRA} with set-based prediction of other obstacles, without incorporating traffic rules in the safety verification.} \vspace{2mm} %
		\label{fig:highenv-ours}
	\end{subfigure}
	\begin{subfigure}{1\columnwidth}
		\centering
		\includesvg[width=\linewidth]{figures/frame_0008-safe-rule.svg}
		\caption{Safe decision-making at time step~$8$ using \textsc{SanDRA} with set-based prediction of other obstacles, incorporating traffic rules in the safety verification.} \vspace{2mm} %
		\label{fig:highenv-ours-rule}
	\end{subfigure}
	\begin{subfigure}{1\columnwidth}
		\centering
		\includesvg[width=\linewidth]{figures/robustness.svg}
		\caption{Robustness evaluation of the simulated trajectories for compliance with the traffic rules.} %
		\label{fig:robustness}
	\end{subfigure}
	\caption{Illustrative comparison in {highway-env}\protect\footnotemark\ under setting \small{\textcircled{\scriptsize{2}}}, in which the past occupancies of vehicles are shown for the previous five time steps.}\vspace{-2mm}
	\label{fig:highway-env}
\end{figure}

\subsubsection{Example Scenario}
An example scenario from highway-env was shown in Fig.~\ref{fig:highenv-ini}, where DiLu failed to account for the dynamic coupling between ego vehicle braking and the distance to the preceding vehicle, resulting in a rear-end crash at simulation step $8$ (cf. Fig.~\ref{fig:highenv-dilu}). This issue was mitigated by \textsc{SanDRA} through set-based prediction of surrounding obstacles, which considered possible emergency braking maneuvers of the preceding vehicle during decision-making. Since the ego vehicle was initially too close to the preceding vehicle (cf.~Fig.~\ref{fig:highenv-ini}), it first executed a series of fail-safe actions to create sufficient space before following the LLM-planned actions (cf.~Figs.~\ref{fig:highenv-ours} and \ref{fig:highenv-ours-rule}). When formalized traffic rules were incorporated in the verification process, this phase was prolonged (cf.~Fig.~\ref{fig:highenv-ours-rule}) to satisfy additional safe-distance requirements in the rule R\_G1 (cf.~(\ref{eq:rg1})).

\footnotetext{Highway-env seed: 5838}
To quantify the extent to which the ego vehicle complied with or violated the three traffic rules in MTL, we additionally presented the robustness \cite[Sec.~VI]{nomoretraffic} of trajectories obtained from the example scenarios in Fig.~\ref{fig:robustness}. A positive robustness value indicated satisfaction of the rules, whereas a negative value indicated violation. The larger the absolute value of the robustness, the stronger the satisfaction or violation of the rules. For a systematic measure of robustness for the general traffic rules in Sec.~\ref{subsec:rule_intro}, we refer the reader to \cite{YuanfeiLinMPR}. Without explicitly considering rules in decision-making, DiLu exhibited a high degree of violations from the very beginning, ultimately leading to a collision. In contrast, safety verification using set-based predictions in \textsc{SanDRA} improved rule compliance of the ego vehicle from the very beginning. Moreover, the variant that incorporated formalized traffic rules achieved the highest robustness in rule compliance on average across scenarios.
%
%

\section{Conclusion}\label{sec:conc}
We present the first framework for safe decision-making in automated vehicles using LLMs. By translating natural language driving actions into LTL$_\text{f}$ formulas, we enable the coupling of set-based reachability analysis with model checking to verify the safety of each action. Unlike existing methods, our approach supports lightweight LLM prompting, addresses inherent safety gaps in LLMs, and provides provable safety guarantees through set-based predictions of surrounding obstacles and the verification of formalized traffic rules. Additionally, our approach bridges decision-making and downstream trajectory planning by leveraging the computed reachable sets as planning constraints. Future work includes exploring the selection of optimal decision-making horizons, employing vision-language models with surround-view images or point clouds as inputs to achieve a broader multimodal understanding of the environment, and deploying our approach on a research vehicle in real-world scenarios.


\section*{Acknowledgment}
The authors gratefully acknowledge the financial support provided by the German Research Foundation (DFG) under grant AL 1185/9-1, as well as the provision of API credits through the OpenAI Research Access Program.
\bibliographystyle{IEEEtran}
{\bibliography{llm}}

\begin{IEEEbiography}[{\includegraphics[width=1in,height=1.25in,clip,keepaspectratio]{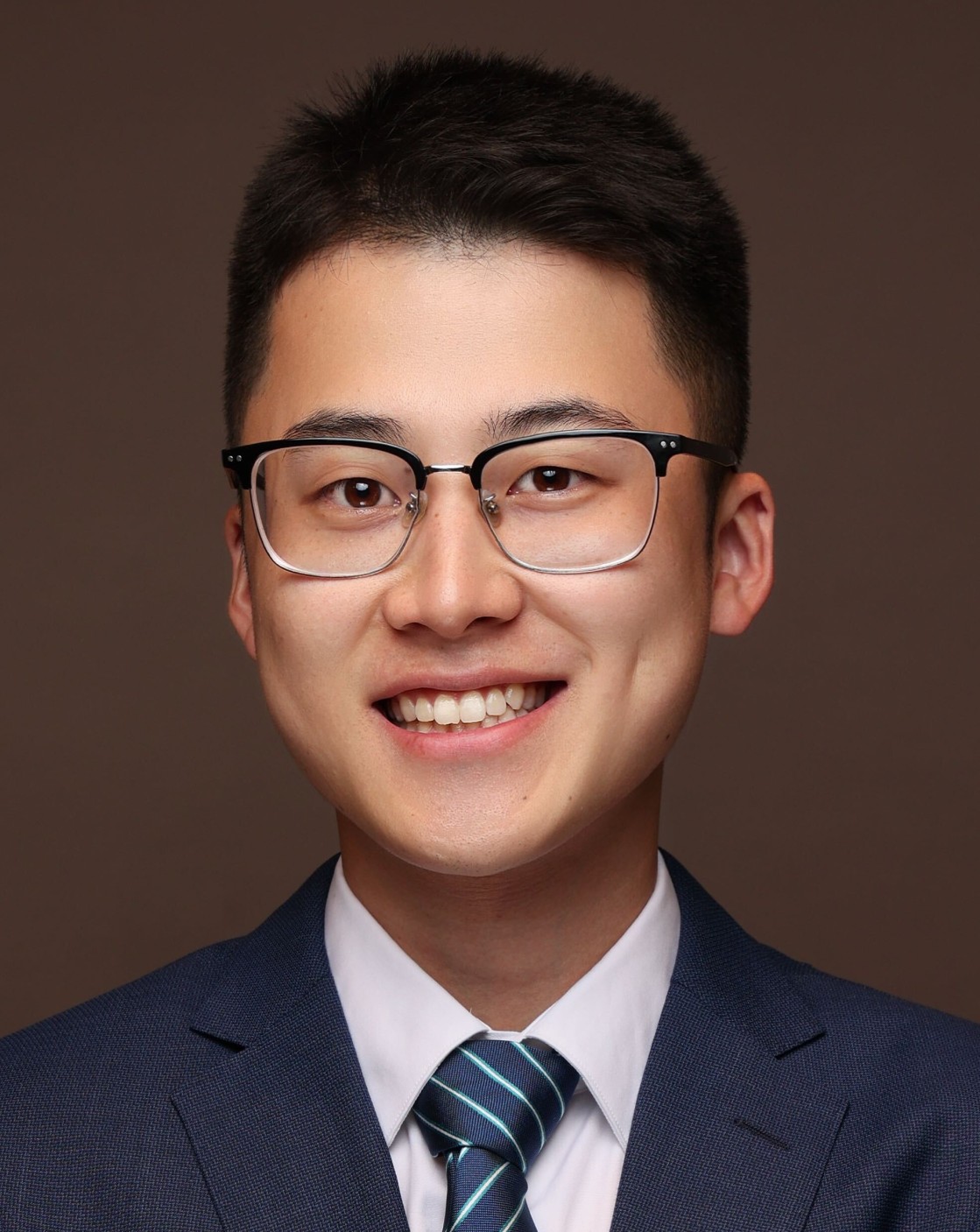}}]{Yuanfei Lin} 
	received his B.Eng. degree in Automotive Engineering from Tongji University, China, in 2018, and dual M.Sc. degrees in Mechanical Engineering and Mechatronics and Robotics at the Technical University of Munich, Germany, in 2020.  In 2023, he was a visiting scholar at the University of California, Berkeley, USA. He completed his Ph.D. in Computer Science at the Technical University of Munich, Germany, in 2025. His research interests include motion planning, formal methods, and large language models for automated vehicles.
\end{IEEEbiography}

\begin{IEEEbiography}[{\includegraphics[width=1in,height=1.25in,clip,keepaspectratio]{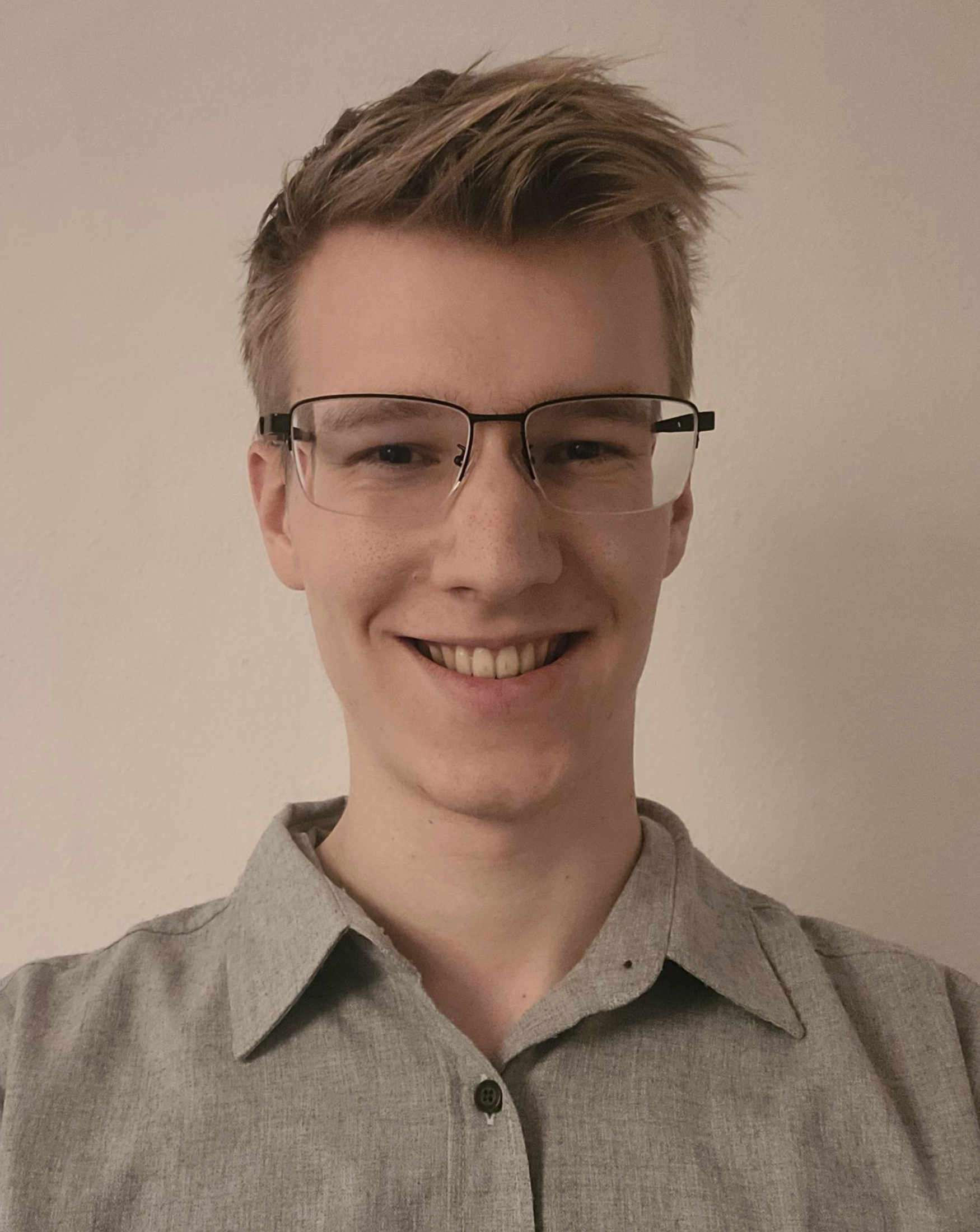}}]{Sebastian Illing} is currently pursuing a M.Sc. degree in Computer Science at the Technical University of Munich, Germany. He received his B.Sc. in Informatics: Games Engineering in 2024, also from the Technical University of Munich, Germany. His research interests include large language models for autonomous driving and 3D Computer Vision.
\end{IEEEbiography}

\begin{IEEEbiography}[{\includegraphics[width=1in,height=1.25in,clip,keepaspectratio]{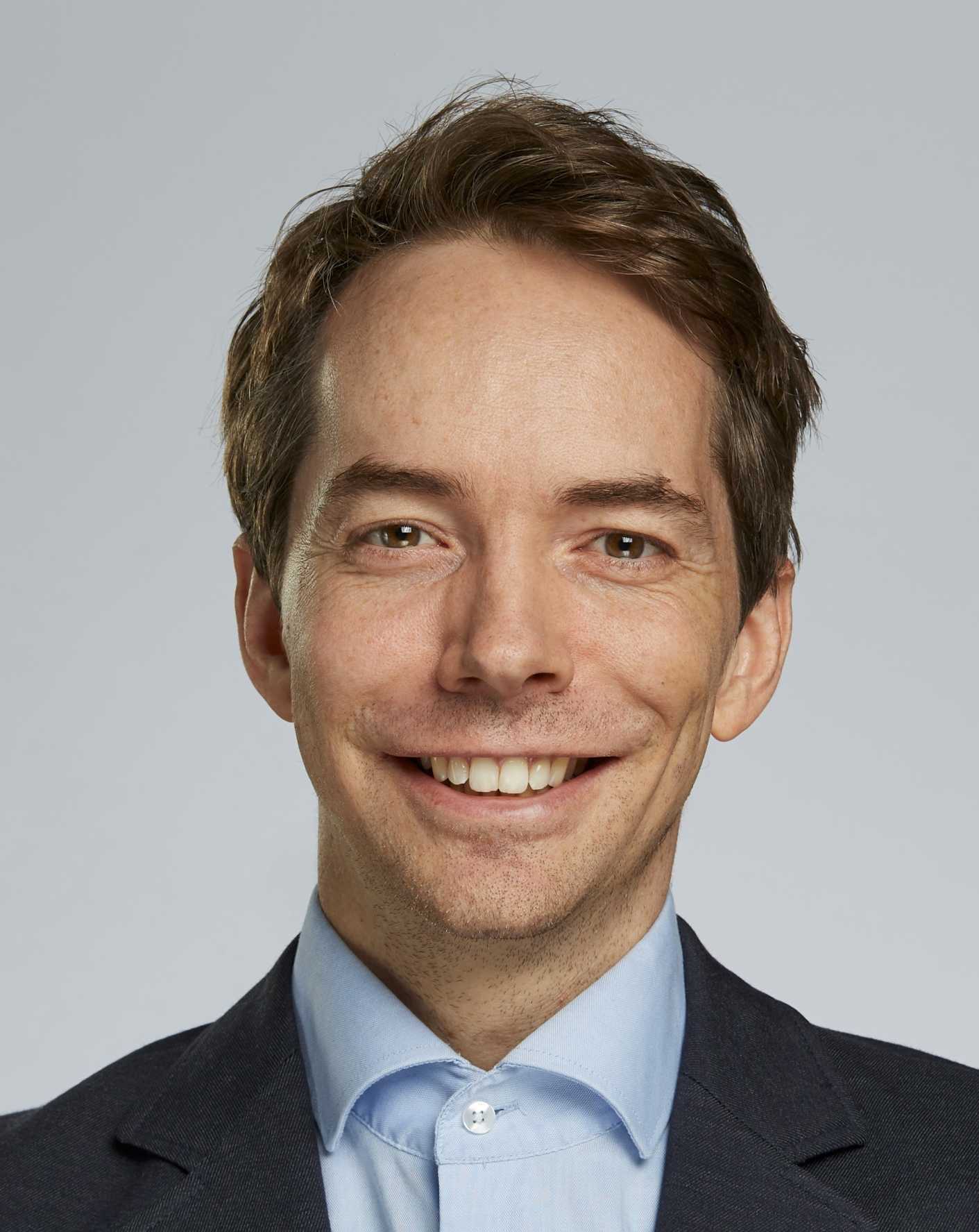}}]
	{Matthias Althoff} is an associate professor in computer science at the Technical University of Munich, Germany. He received his diploma engineering degree in Mechanical
	Engineering in 2005, and his Ph.D. degree in Electrical Engineering in
	2010, both from the Technical University of Munich, Germany.
	From 2010 to 2012 he was a postdoctoral researcher at Carnegie Mellon University,
	Pittsburgh, USA, and from 2012 to 2013 an assistant professor at Technische Universit\"at Ilmenau, Germany. His research interests include formal verification of continuous and hybrid systems, reachability analysis, planning algorithms, nonlinear control, automated vehicles, and power systems.
\end{IEEEbiography}

\vfill

\end{document}